\documentclass{bmvc2k}

\usepackage{graphicx}
\usepackage{booktabs}
\usepackage{rotating}
\usepackage{multirow}
\usepackage{makecell}
\usepackage{float}

\definecolor{highlightcolor}{rgb}{0.0, 0.5, 0.0}
\newcommand{\blue}[1]{\textcolor{bmv@captioncolor}{#1}}

\title{Into the Fog: Evaluating Robustness of\\ Multiple Object Tracking}
\addauthor{Nadezda Kirillova}{nadezda.kirillova@tugraz.at}{1}
\addauthor{M. Jehanzeb Mirza}{mirza@tugraz.at}{1}
\addauthor{Horst Bischof}{horst.bischof@tugraz.at}{1}
\addauthor{Horst Possegger}{possegger@tugraz.at}{1}
\addinstitution{
Institute of Computer Graphics and Vision\\
Graz University of Technology\\
Graz, Austria
}
\runninghead{Kirillova \etal}{Into the Fog: Evaluating MOT Robustness}

\def\etal{\emph{et al}\bmvaOneDot}

\begin{document}
\maketitle

\begin{abstract}
State-of-the-art Multiple Object Tracking (MOT) approaches have shown remarkable performance when trained and evaluated on current benchmarks.
However, these benchmarks primarily consist of clear weather scenarios, overlooking adverse atmospheric conditions such as fog, haze, smoke and dust. As a result, the robustness of trackers against these challenging conditions remains underexplored. 
To address this gap, we introduce 
physics-based volumetric fog simulation method for arbitrary MOT datasets, utilizing frame-by-frame monocular depth estimation and a fog formation optical model. 
We enhance our simulation by rendering both homogeneous and heterogeneous fog and propose to use the dark channel prior method to estimate atmospheric light, showing promising results even in night and indoor scenes.
We present the leading benchmark MOTChallenge (third release) augmented with fog (smoke for indoor scenes) of various intensities and conduct a comprehensive evaluation of MOT methods, revealing their limitations under fog and fog-like~challenges.
\end{abstract}

\section{Introduction}
\label{sec:intro}
Multiple object tracking (MOT) plays a core role in providing situation awareness for intelligent video surveillance systems, smart city technologies and autonomous driving. 
Driven by advancements in computer vision and deep learning, MOT methods have achieved remarkable results when trained and evaluated on current benchmarks~\cite{motsurvey, motchallenge, KITTI, nuscenes, waymo}.
However, these benchmarks primarily consist of clear weather scenarios,
overlooking performance degradation under adverse atmospheric conditions, including fog and fog-like phenomena such as haze, dust, and smoke.
To the best of our knowledge, no comprehensive work is available that extensively analyzes the robustness of MOT methods to such challenges.
Moreover, only a limited number of appropriate datasets exist \cite{rainrendersite, motsynth, nuscenesfog}, which are, however, still insufficient and do not cover the full range of fog intensity.

Collecting and annotating new data is time-consuming and labor-intensive, especially under adverse atmospheric conditions.
Moreover, data protection laws in many countries may raise privacy concerns due to potential collection of sensitive data.
To overcome these challenges, we utilize existing clear MOT datasets and generate photorealistic fog (smoke for indoor scenes) of various intensity levels.
The approach of simulating adverse conditions in real-world clear images has been successfully demonstrated in various computer vision tasks such as image classification~\cite{3dcommoncorruptions, commoncorruptions, augmix, albumentations}, object detection~\cite{rainrender, nuscenesfog} and semantic segmentation~\cite{foggycityscape}. 
However, its application to the MOT domain has remained unexplored. 

\begin{figure}
\centering
  \resizebox{\linewidth}{!}{
  \rotatebox{90}{\scriptsize \hspace{7pt} MOT17-02}
  \includegraphics[width=0.25\linewidth]{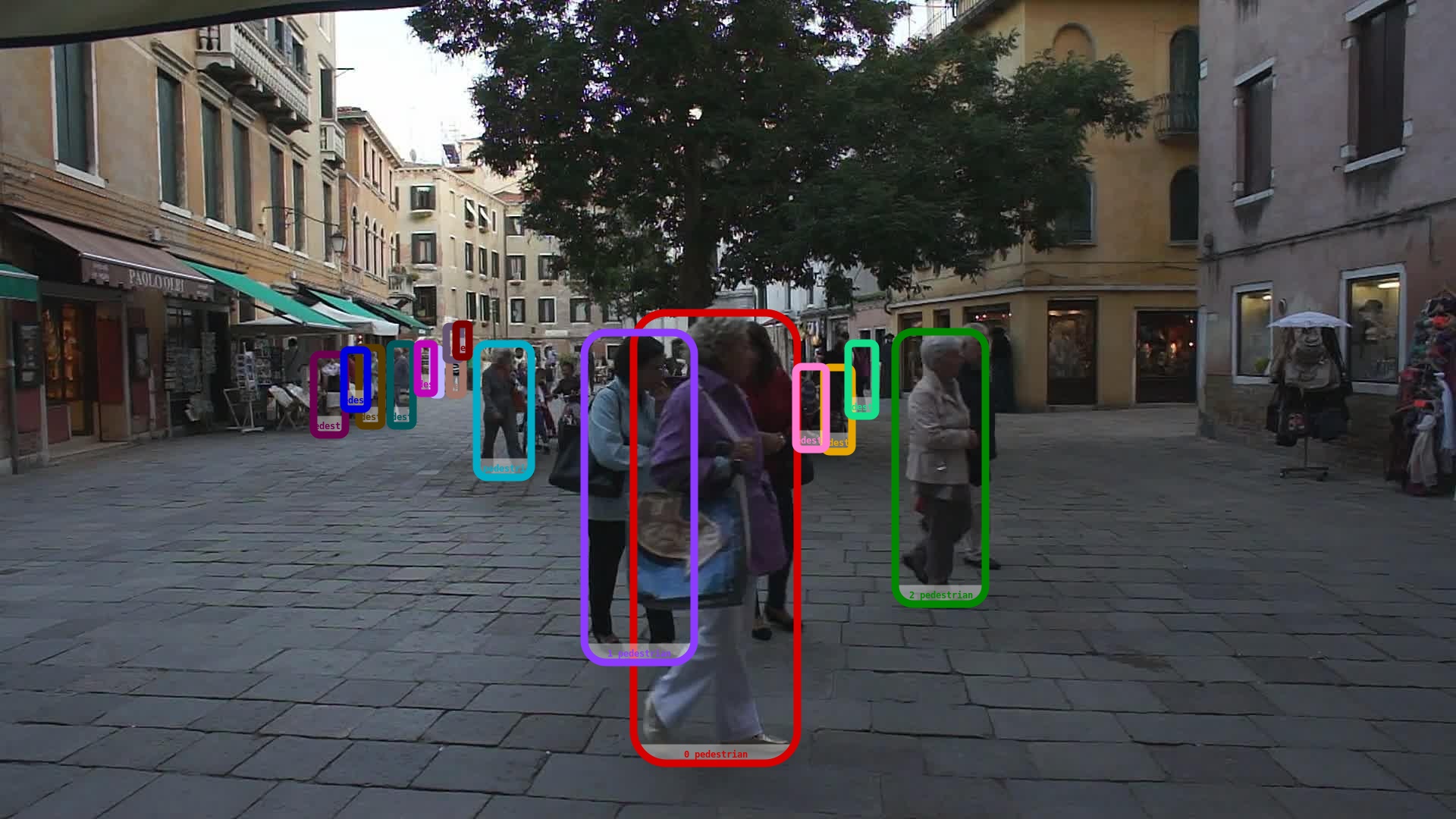}%
  \includegraphics[width=0.25\linewidth]{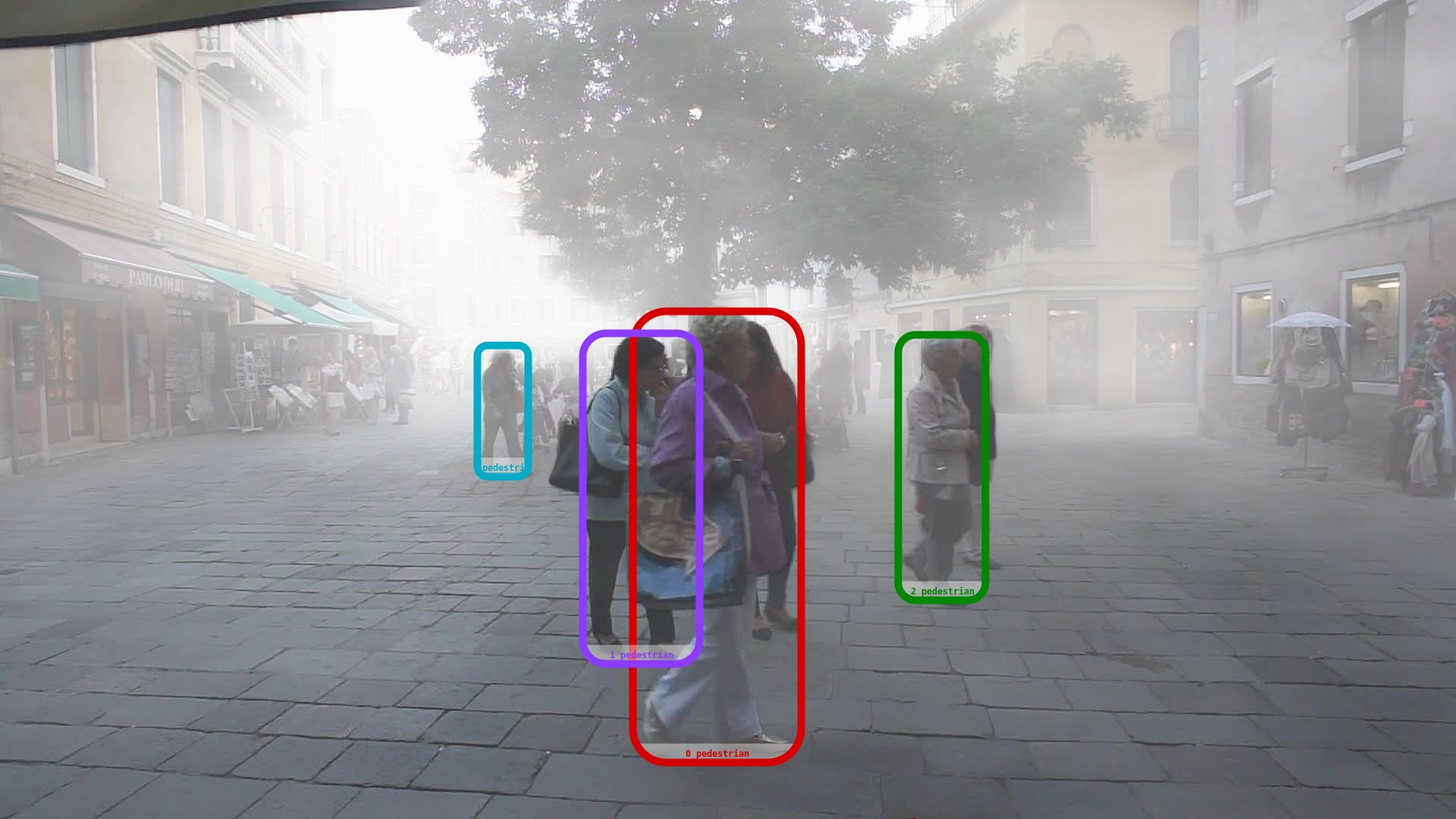}%
  \includegraphics[width=0.25\linewidth]{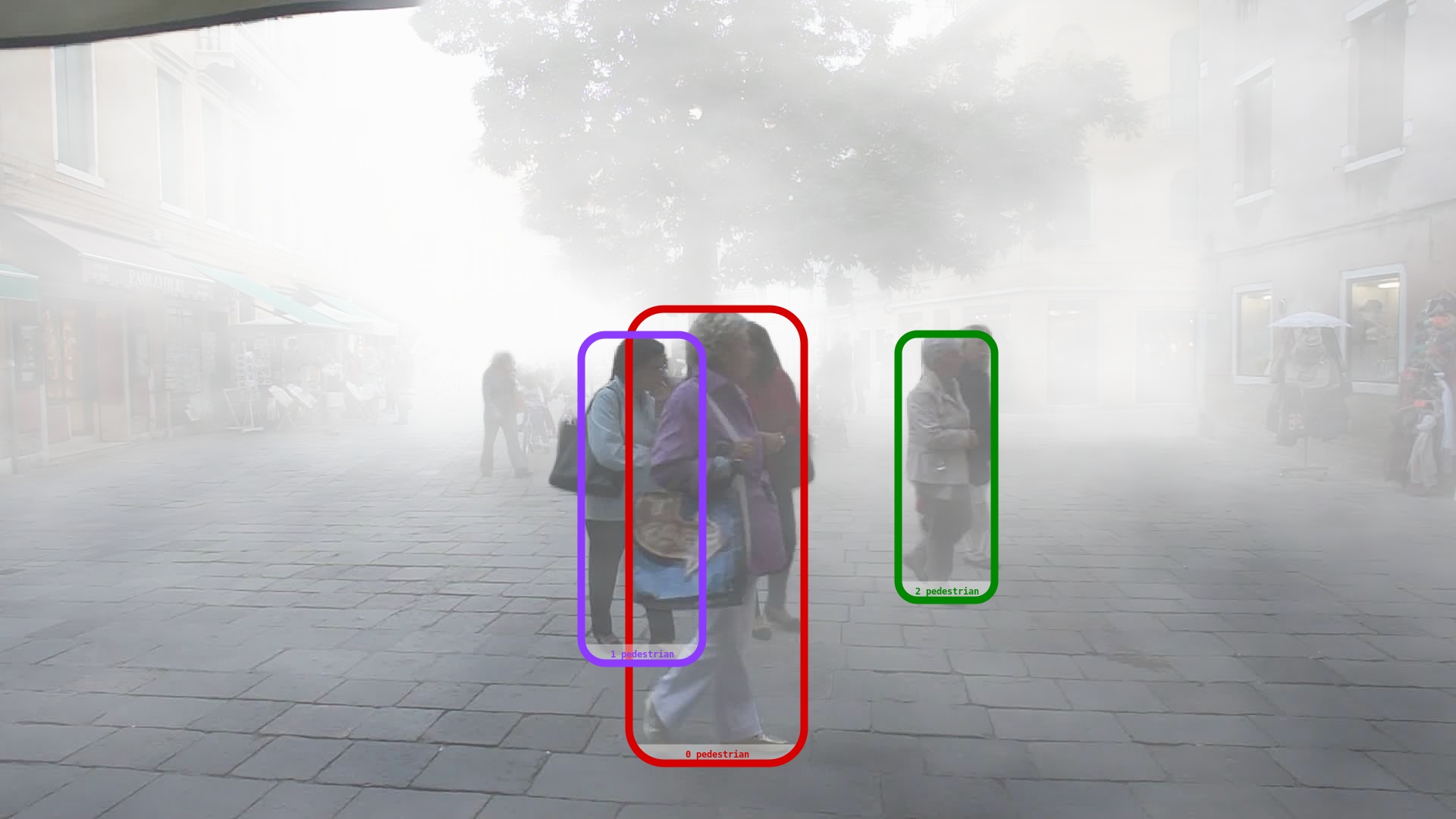}%
  \includegraphics[width=0.25\linewidth]{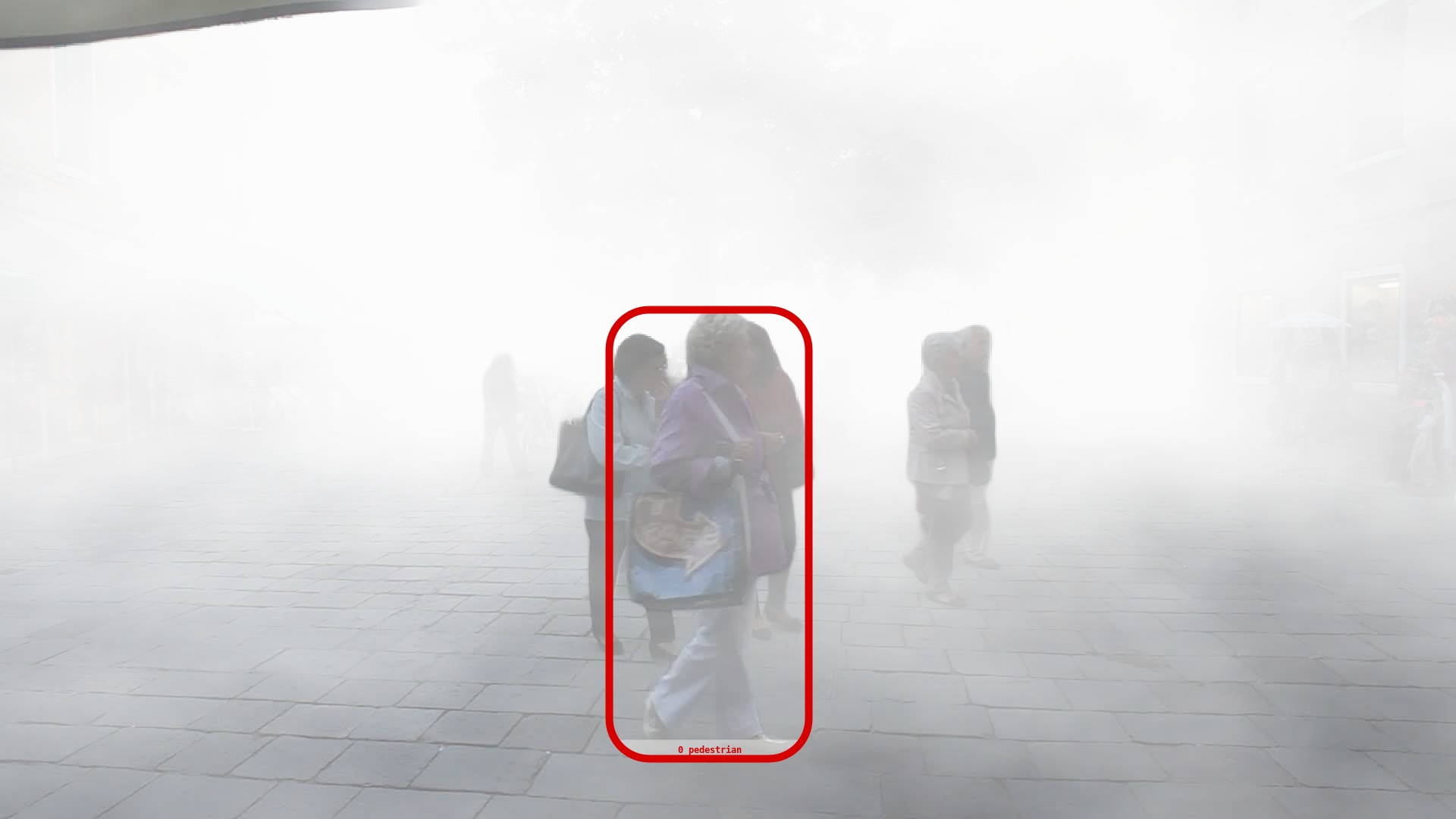}}\\
  \resizebox{\linewidth}{!}{
  \rotatebox{90}{\scriptsize \hspace{7pt} MOT17-04}
  \includegraphics[width=0.25\linewidth]{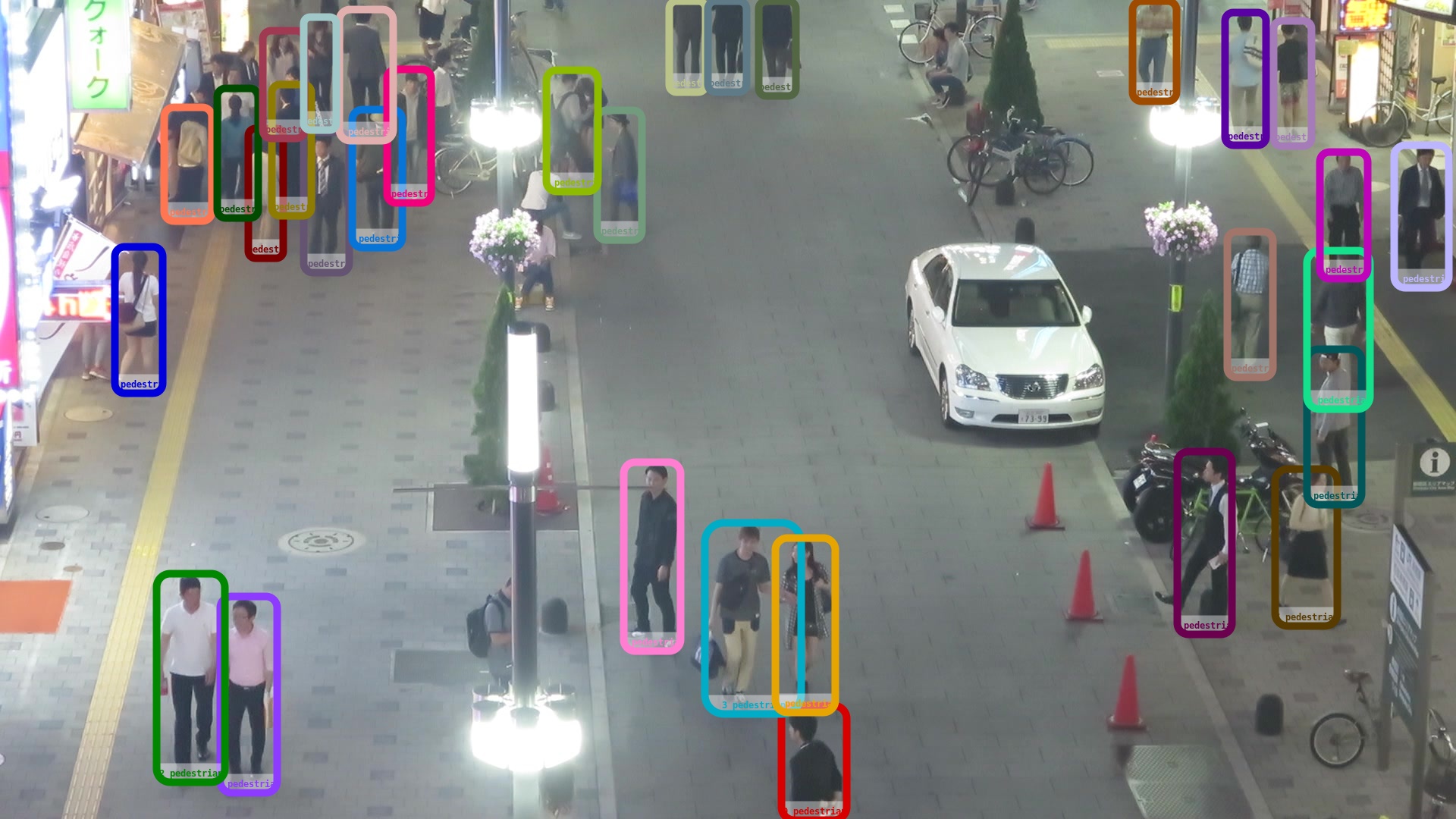}%
  \includegraphics[width=0.25\linewidth]{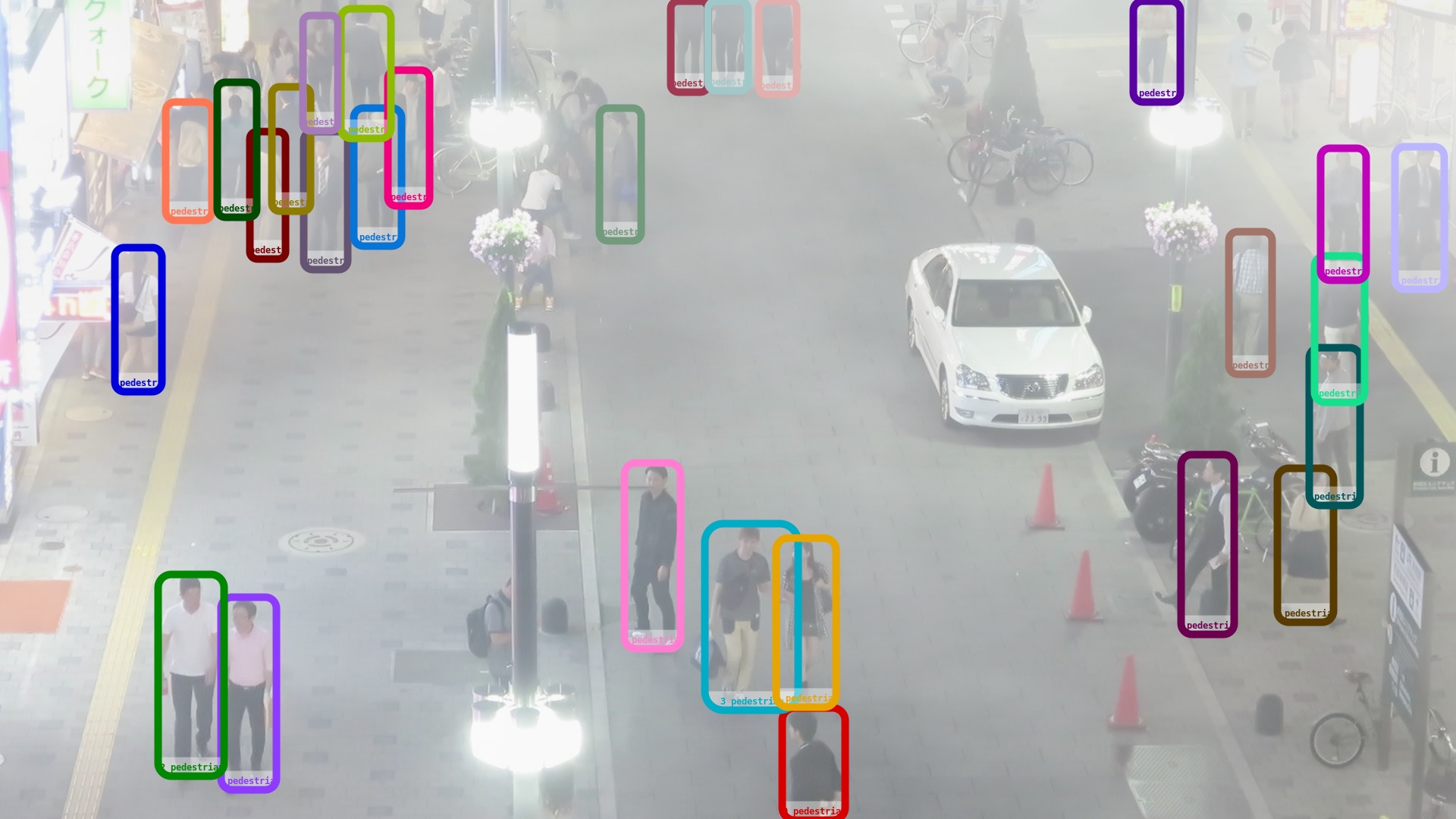}%
  \includegraphics[width=0.25\linewidth]{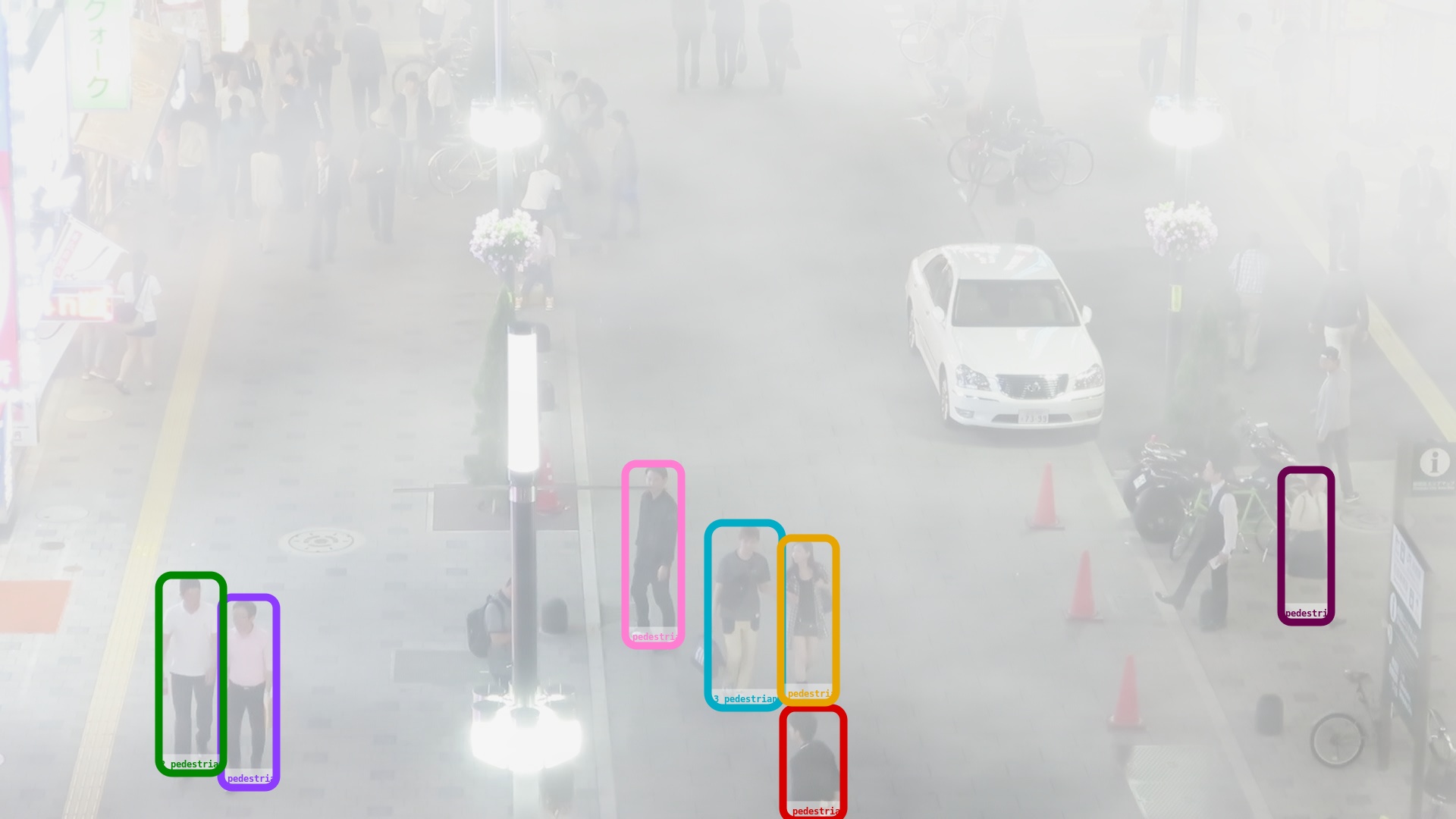}%
  \includegraphics[width=0.25\linewidth]{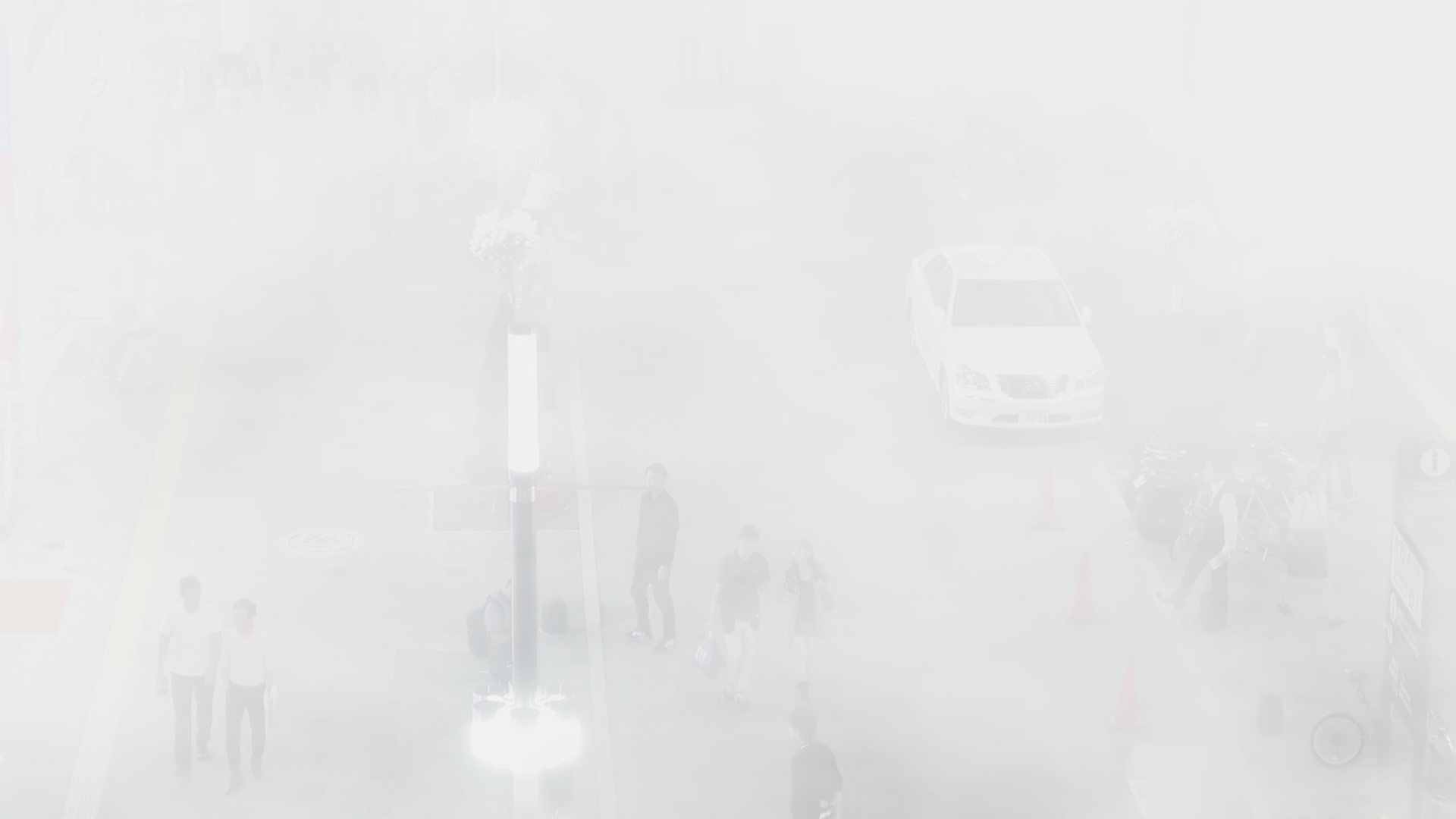}}\\
  \caption{Visualization of MOT performance degradation with increasing fog intensity (from left to right), demonstrated on fog-augmented video sequences with different scenarios. Top row: a daytime front-view static camera scene. Bottom row: a heavily illuminated night crowd surveillance scene. (Best viewed on screen.) }
  \label{fig:main}
  \vspace{-5pt}
\end{figure}

Our work presents a novel extension of volumetric fog rendering to the MOT task, which, unlike the previously mentioned tasks, operates on videos rather than single images.
In comparison with previous methods~\cite{foggycityscape, rainrendersite, nuscenesfog}, which apply volumetric fog rendering, we do not rely solely on autonomous driving (AD) datasets, where accurate depth information is provided by range sensors, such as LiDAR, stereo camera or ToF.
We introduce a novel twist by leveraging monocular depth estimation.
This adaptation allows us to extend the feasibility of volumetric fog simulation beyond the scope of AD, opening up new possibilities 
across diverse scenarios.
Additionally, we enhance our simulation by rendering of both homogeneous and heterogeneous fog effects to achieve more diversity in the representation of this complex natural phenomenon.
Unlike previous studies~\cite{foggycityscape, rainrendersite, nuscenesfog}, our approach is not restricted to images where the sky is visible. This flexibility stems from leveraging the dark channel prior method~\cite{He} for atmospheric light estimation and subsequent fog color determination, yielding promising results even in night and indoor scenes.
Finally, we augment with fog (smoke for indoor scenes) of various intensities the leading tracking benchmark MOTChallenge~(third release: MOT17 dataset)~\cite{motchallenge}, which represents the combination of datasets from different domains and known for its diversity and wide coverage of various real-world scenarios. 
We conduct a comprehensive evaluation of different state-of-the-art (SOTA) MOT methods, quantitatively analyzing the impact of fog and fog-like phenomena on MOT robustness, thereby revealing their limitations, as illustrated in Fig.~\ref{fig:main}.

\section{Related Work}
Relevant work to our research involves studies on MOT robustness, availability of appropriate benchmarks, semi-synthetic simulations in real-world images and methods for extracting depth information from a 2D scene.

\subsection{MOT Robustness} 
MOT is a complex computer vision task that includes various steps, such as object detection, motion estimation, instance re-identification (re-ID) and data association. 
MOT deals with video sequences rather than separate images, incorporating the temporal information.
Frame-by-frame, it processes visual data from the surrounding environment, detects and tracks targets over time, forming their trajectories.
Abundant SOTA methods~\cite{motsurvey, centertrack, FairMOT, bytetrack, tracktor, SORT, DeepSort} suggest different approaches to address various MOT challenges such as occlusion, viewpoint variation, pose modification, appearance changes, scale diversity, crowded scenes, lighting conditions, real-time processing, and model complexity. 
However, research on MOT robustness in adverse atmospheric conditions is still lacking.
Some early works~\cite{tan, He, defogging, restoration} propose general dehazing techniques to address issues caused by fog or haze by restoring clear scenes.
However, image restoration approaches face limitations in model bias and generalization ability, loss of image details or textures, and misinterpretation of scene ambiguity.
The problem of developing MOT methods capable to handle adverse atmospheric conditions, including fog, remains relevant and underexplored.

\subsection{Adverse Weather Datasets}
Availability of datasets with adverse weather conditions is crucial for improving MOT model robustness.
While several works release such datasets (see Tab.~\ref{tab:weathdatasets}), especially in the AD domain, they still face limitations and need more diversity. 
The real-world BDD100K~\cite{BDD100k} dataset, recorded from car dashcams, includes video footage of adverse weather conditions but suffers from issues, such as camera shaking, poor image quality, and does not contain foggy videos. 
The DAWN~\cite{dawn} dataset, composed from web sources, offers diverse images of fog, rain, snow and sand, but lacks video sequences necessary for tracking purposes. 
Among datasets for autonomous driving (AD), such as KITTI~\cite{KITTI}, NuScenes~\cite{nuscenes}, Waymo Open~\cite{waymo}, Cityscapes~\cite{Cityscapes}, Oxford RobotCar~\cite{RobotCar}, CADC~\cite{CADC} and DENSE~\cite{dense}, only the last one has foggy images. 
While all of them provide range data sufficient for fog rendering, only the first three offer MOT annotations, and they all are constrained to specific camera positions typical for the AD domain.
Fully synthetic FRIDA~\cite{FRIDA} and FRIDA2~\cite{FRIDA2} datasets, which provide foggy scenes, are also captured from the driver's perspective. They are designed to address dehazing problems and do not contain videos, which makes them unsuitable for MOT tasks.
CARLA~\cite{carla} is a constantly growing open-source simulator for AD research that offers a realistic synthetic environment for testing and developing AD models across various scenarios, including foggy and rainy conditions. 
However, the fully synthetic Omni-MOT dataset~\cite{omnimot}, which was derived from CARLA and created for tracking purposes, does not include foggy scenarios.
The MOTSynth~\cite{motsynth} dataset, generated from the Grand Theft Auto V (GTA-V) computer game, aims to address tracking problems in various scenarios but lacks variation in fog intensity and scene diversity. Moreover, MOTSynth is fully synthetic, that only widens the gap with reality compared to our semi-synthetic approach.

The MOTChallenge~\cite{motchallenge} benchmark is a leading platform for evaluating MOT methods and video scene understanding. 
Its third release, the MOT17 dataset, covers a wide range of data diversity across different domains, including scenes with moving and static cameras, captured indoors and outdoors from different camera positions, as well as scenarios from surveillance and autonomous driving, and spanning from day to night. 
In this work, we demonstrate our method for photorealistic fog simulation using the MOTChallenge benchmark, as it provides the most general combination of different MOT dataset types.

\subsection{Semi-synthetic Simulations in Real-world Images}
Efforts to create semi-synthetic images based on real ones have been undertaken across various domains and shown promising results in improving the robustness of deep learning models. 
In the image classification task, some works~\cite{commoncorruptions, augmix, albumentations} propose common image corruption techniques for \textsc{ImageNet}~\cite{imagenet} and CIFAR~\cite{cifar}, paving the way to enhance model generalization.
However, they involve non-realistic planar fog simulations, diverging from real-world scenarios. 
To overcome these limitations, Kar~\etal~\cite{3dcommoncorruptions} propose a monocular depth estimation approach to integrate scene geometry for more realistic simulations, demonstrating its effectiveness.
However, these corruptions are constrained to homogeneous volumetric fog (smoke) in indoor single images captured from a front-view camera position. 

Sakaridis~\etal~\cite{foggycityscape} explore the fog impact on semantic segmentation performance in the AD domain. They generate volumetric fog in real-world single images and collect a small foggy dataset to validate the approach. 
However, their study is limited to the Cityscapes~\cite{Cityscapes} dataset and relies on provided stereo images along with corresponding disparity maps. 
Moreover, their fog simulation is only feasible in images containing the sky and is restricted by homogeneity.
Similarly, some other AD works with available range data~\cite{rainrender, nuscenesfog} encounter such limitations, focusing on fog simulation in \textsc{KITTI}~\cite{KITTI} and NuScenes~\cite{nuscenes} datasets with the aim to enhance the object detection task. 
Despite the promising results of employing synthetic fog in these contexts, there are currently no physics-based volumetric fog simulations available to address MOT requirements with arbitrary scenarios and camera viewpoints.

\subsection{Monocular Depth Estimation}
Photorealistic volumetric fog simulation requires 3D scene information and relies on image depth maps. 
Accurate monocular depth estimation (MDE) remains an ongoing research challenge due to the inherent uncertainty in reconstructing the 3D structure of a scene from a single 2D image. 
MDE faces various challenges, including scale ambiguity and generalization ability across diverse domains.
Moreover, processing videos on a per-frame basis results in temporal inconsistency for dynamic scenes, leading to fog flickering over time~\cite{fog4game, temporalconsistent}. 
Recent MDE methods \cite{midas, dpt, DepthAnything, Zoe} use vision transformers~\cite{transformers}, which, compared to convolutional neural networks, exhibit higher accuracy.
One line of these works~\cite{Zoe, DepthAnything} focuses on \textit{metric depth} estimation directly from a model. However, they are predominantly fine-tuned on specific narrow datasets, resulting in overfitting and encountering generalization limitations.
To facilitate effective training across multiple datasets and enhance generalization across different domains, \textit{relative depth} estimation approaches~\cite{midas, dpt, relative_depth} are employed.
Such methods predict pixel-wise disparity without providing metric values and allow us to utilize MDE on arbitrary MOT sequences, where fine-tuning is not possible.

\begin{table*}[th]
  \centering
  \resizebox{\textwidth}{!}{\begin{tabular}{@{}clcccccc@{}}
    \toprule
     & Dataset & Year & Weather & Task & Camera View & 3D Info & Augmentation \\
    \midrule
    \multirow{17}{*}{
    \begin{sideways}
        Real-world data
    \end{sideways}}
            
            & KITTI~\cite{KITTI} & 2014 & clear & \makecell{D2D, \blue{MOT2D}, DE, LD,\\ OF, SG, SLAM, ST}& AD & \makecell{depth map,\\ range data} & \makecell{\blue{fog}~\cite{rainrendersite}, rain~\cite{rainrendersite}}\\
            \cmidrule{3-8}
                        
            & \makecell[l]{MOTChallenge~\cite{motchallenge,mot20}} & \makecell{2015-20}&clear & \makecell{\blue{MOT2D}, VSU} & diverse & - & - \\
            \cmidrule{3-8}
            
            & Cityscapes~\cite{Cityscapes}& 2016 & clear & \makecell{D3D, SG}& driver & \makecell{depth map\\ from stereo} & \makecell{\blue{fog}~\cite{foggycityscape, rainrendersite}, rain~\cite{rainrender}}\\
            \cmidrule{3-8}
                    
            & BDD100K~\cite{BDD100k} & 2017 & \makecell{clear,\\ rain, snow, fog$^*$} & \makecell{D2D, \blue{MOT2D}, \\LD, MOTS, SG}& dashcam & - & -\\
            \cmidrule{3-8}
             
            & \makecell[l]{Oxford RobotCar~\cite{RobotCar}} & 2017 &\makecell{clear, rain, snow} & SLAM & AD & range data& - \\
            \cmidrule{3-8}
  
            & NuScenes~\cite{nuscenes} & 2019 &\makecell{clear, rain } & \makecell{D3D, \blue{MOT3D}, SLAM, MP}& AD & \makecell{range data} & \makecell{\blue{fog}~\cite{nuscenesfog}, rain~\cite{rainrender}}\\
            \cmidrule{3-8}

            & \makecell[l]{Waymo Open~\cite{waymo}} & 2019 &\makecell{clear, rain} & \makecell{D2D, \blue{MOT2D}, D3D, MOT3D,\\DA, HPE, SG, SLAM, MP} & AD & range data & -\\
            \cmidrule{3-8}
                                   
            & CADC~\cite{CADC} & 2020 &\makecell{snow (variety)}  & \makecell{D3D, SLAM}& AD & range data & -\\
            \cmidrule{3-8}
  
            & DAWN~\cite{dawn} & 2020 &\makecell{\blue{fog}, rain,\\sand, snow} & D2D & \makecell{web} & - &-\\
            \cmidrule{3-8}

            & DENSE~\cite{dense} & 2020 &\makecell{clear, rain, snow,\\ \blue{fog} chamber} & \makecell{D2D, D3D, \\DA, DE, MSF} & AD & \makecell{depth map,\\range data} & - \\
    \midrule
    \multirow{8}{*}{
    \begin{sideways}
       Synthetic data
    \end{sideways}} 

            & \makecell[l]{FRIDA~\cite{FRIDA}\\FRIDA2~\cite{FRIDA2}} & \makecell{2010\\2012} & \makecell{clear, \blue{fog}} & \makecell{dehazing} & driver & \makecell{depth map,\\calibration} &  \\
            \cmidrule{3-8}
            
            & CARLA~\cite{carla} & 2017 &\makecell{clear, \blue{fog}, rain} & \makecell{driving simulator} & \makecell{virtual AD\\multi-cam.} & \makecell{depth map,\\range data} & \\
            \cmidrule{3-8}

            & Omni-MOT~\cite{omnimot} & 2020 &\makecell{clear, rain} & \blue{MOT2D} & \makecell{CARLA} & \makecell{velocity,\\calibration} &  \\  
            \cmidrule{3-8}            

            & MOTSynth~\cite{motsynth} & 2021 &\makecell{clear, \blue{fog}, rain \\ snow, thunder} & \makecell{\blue{MOT2D}, MOTS, VSU} & GTA-V & depth map &  \\
    \bottomrule
  \end{tabular}}
  \caption{Overview of datasets providing either adverse weather, its simulation, or MOT annotations, with the following task abbreviations:  
      \textit{D2D} (\textit{D3D}): 2D (3D) object~detection;
      \textit{DA}: domain adaptation;
      \textit{DE}: depth estimation;
      \textit{HPE}: human pose estimation;
      \textit{LD}: lane detection;
      \textit{MOT2D} (\textit{MOT3D}): 2D (3D) MOT;
      \textit{MOTS}: MOT and segmentation; 
      \textit{MP}: motion prediction;
      \textit{MSF}: multi-sensor fusion;      
      \textit{OF}: optical flow; 
      \textit{SG}: segmentation; 
      \textit{SLAM}: simultaneous localization and mapping;
      \textit{ST}: stereo evaluation;
      \textit{VSU}: video scene understanding.\\
  \begin{footnotesize}      
  $^{*}$only one foggy sequence is available but was incorrectly annotated (is actually \textit{clear}).
  \end{footnotesize}
  }
  \vspace{-7pt}
  \label{tab:weathdatasets}
\end{table*}

\section{Volumetric Fog Rendering}
Depending on various factors such as humidity, temperature and air movement, natural fog varies in density and thickness, significantly affecting the visibility of distant objects~\cite{visionatmospher, Middleton1951,McCartney}. Accurately depicting such a complex phenomenon is a challenging task that must consider many factors.  
This section describes our framework for physic-based volumetric fog rendering applicable to an arbitrary MOT dataset.

\subsection{Depth Maps}
\label{seq:depthmap}
Since we aim to render fog in monocular videos without any range data provided, we employ MDE method on a per-frame basis.
We utilize the top-performing MiDaS~3.0~\cite{midas} method with a powerful vision transformer~\cite{transformers} backbone $\text{DPT BEiT}_\text{512}\text{-Large}$~\cite{dpt, beit}. 
This model has robust generalization capabilities across diverse domains due to its zero-shot cross-dataset transfer ability and two mixing training strategies. 
MiDaS achieves strong results in single-image \textit{relative inverse depth} $\mathbf{d(x)}$ estimation (at pixel $\mathbf{x}=(x,y)$), with accuracy up to scale $s$ and shift~$t$.
If 3D ground truth reference points of the captured scene are available, the actual \textit{metric depth} $\mathbf{D(x)}$ can be obtained.
(Details are provided in the supplementary material.)
This allows us to align fog intensity simulations with meteorological \textit{visibility} measured in meters.
We denote fog severity levels ranging from $1$ to $k=4$, where we apply visibility thresholds of less than $100\;m,\; 50\;m,\; 20\;m$ and $10\;m$ for outdoor scenes, and $20\;m,\; 10\;m,\; 5\;m$ and $3\;m$ for indoor scenes, respectively. 
Without 3D reference points, approximate metric depth conversion still enables effective fog simulation. 
However, the fog’s severity may not precisely correspond to the actual visibility.

\begin{figure*}[ht]
  \centering
  \renewcommand{\tabcolsep}{0pt}
  \begin{tabular}{cccc}

    \subfigure[MOT17-02]{\bmvaHangBox{
        \includegraphics[width=0.25\linewidth]{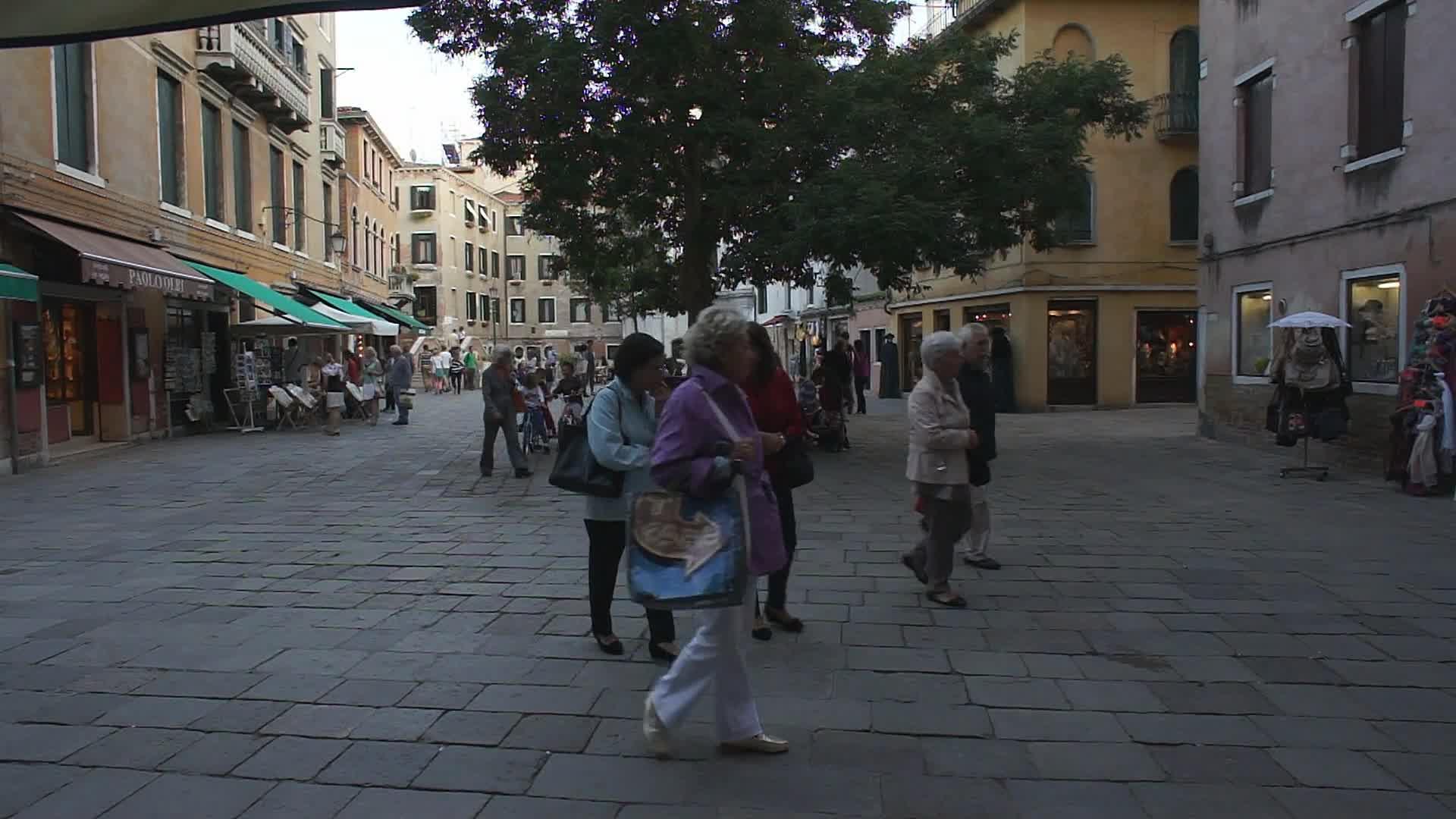}\\[-\baselineskip]
        \includegraphics[width=0.25\linewidth]{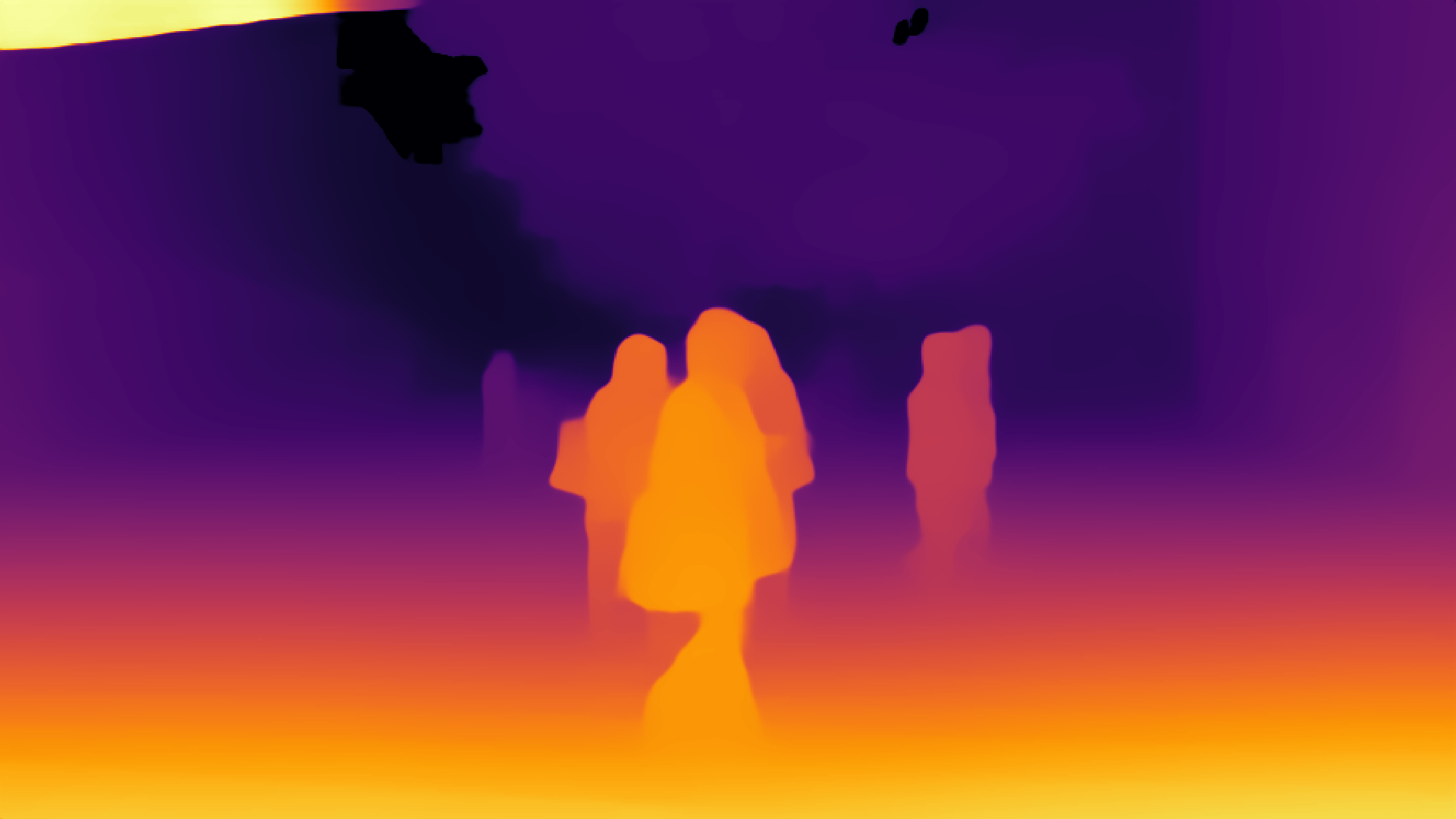}\\[-\baselineskip]
        \includegraphics[width=0.25\linewidth]{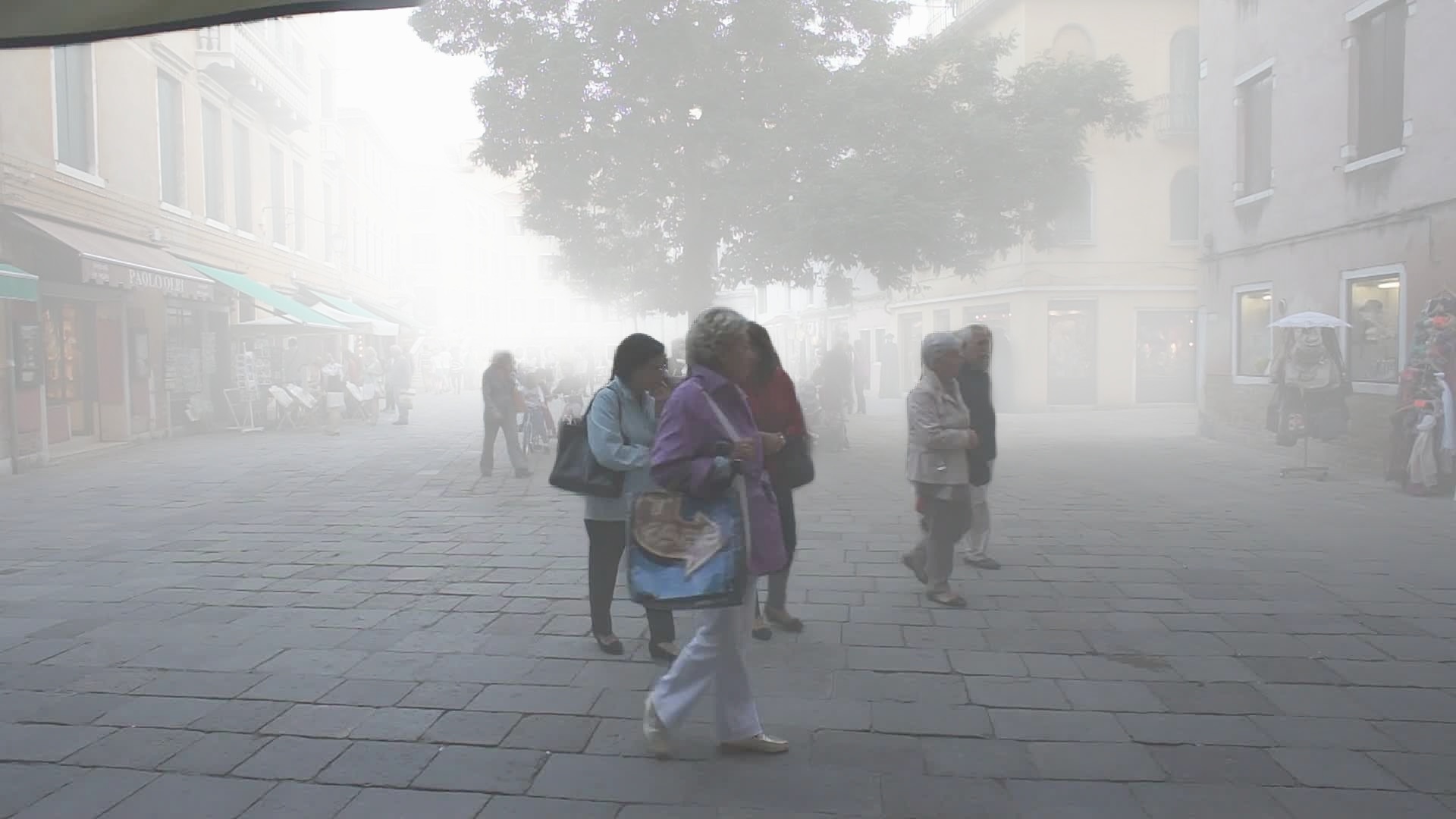}\\[-\baselineskip]}}
    &
    \subfigure[MOT17-04]{\bmvaHangBox{
        \includegraphics[width=0.25\linewidth]{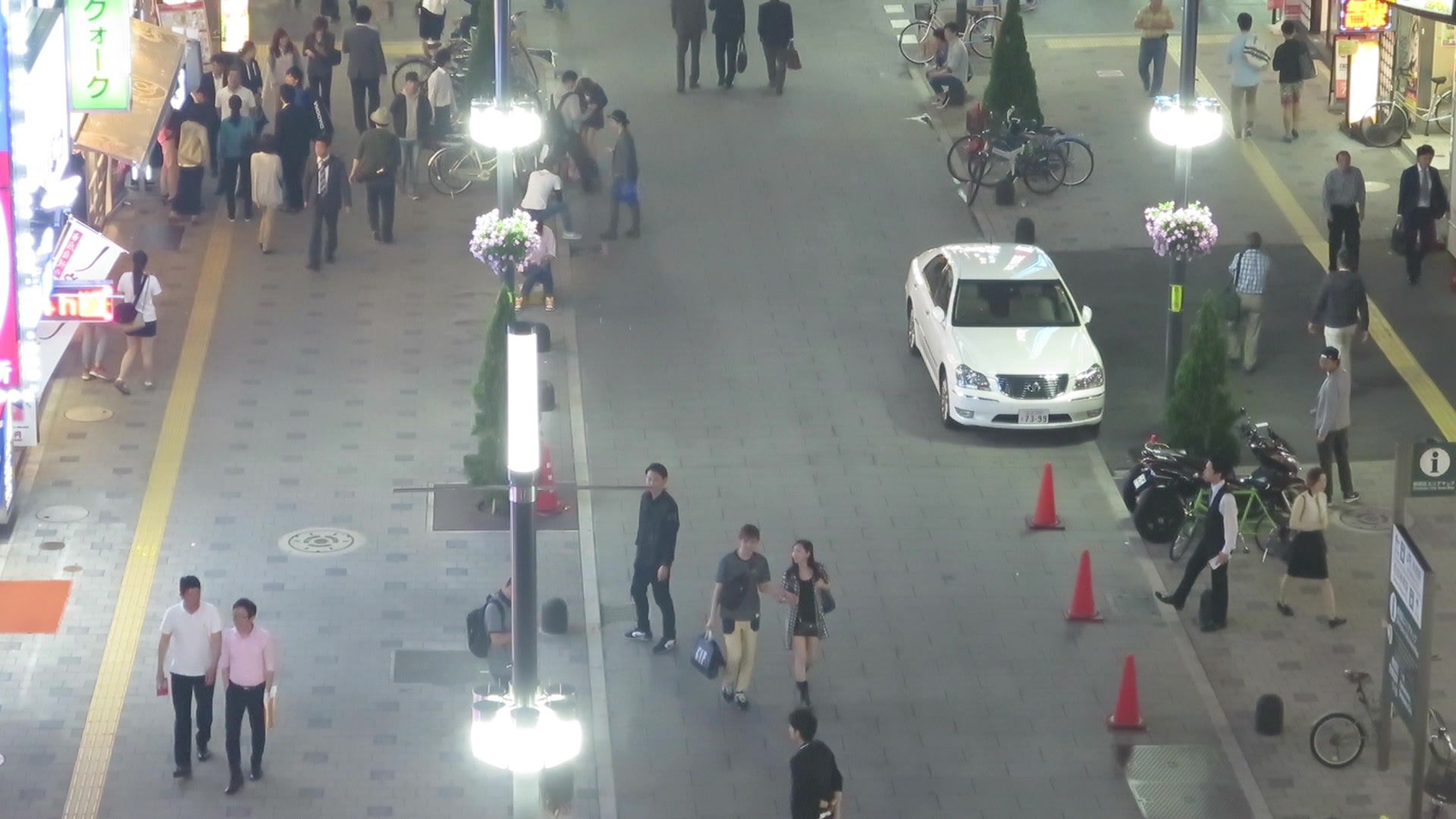}\\[-\baselineskip]
        \includegraphics[width=0.25\linewidth]{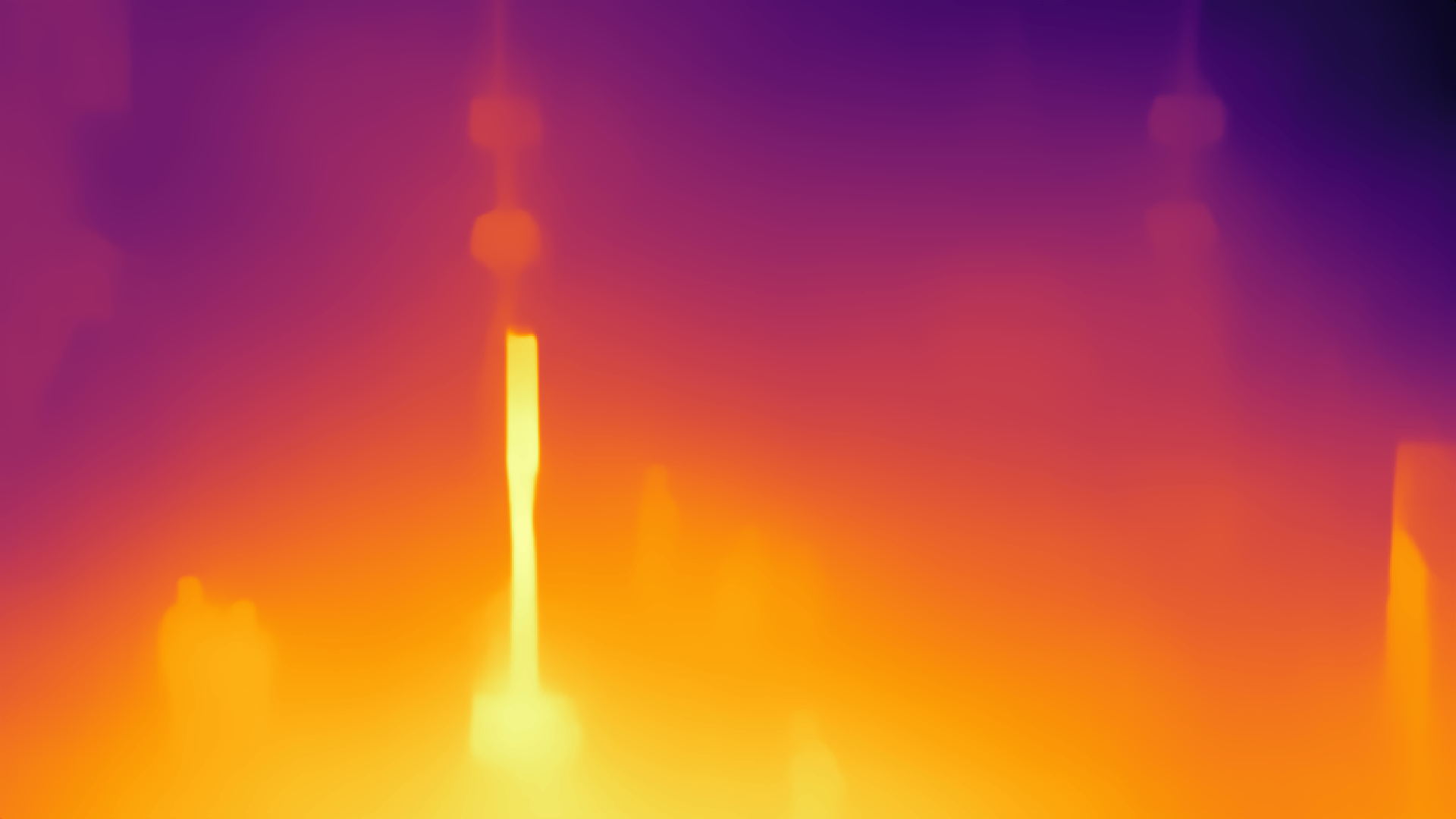}\\[-\baselineskip]
        \includegraphics[width=0.25\linewidth]{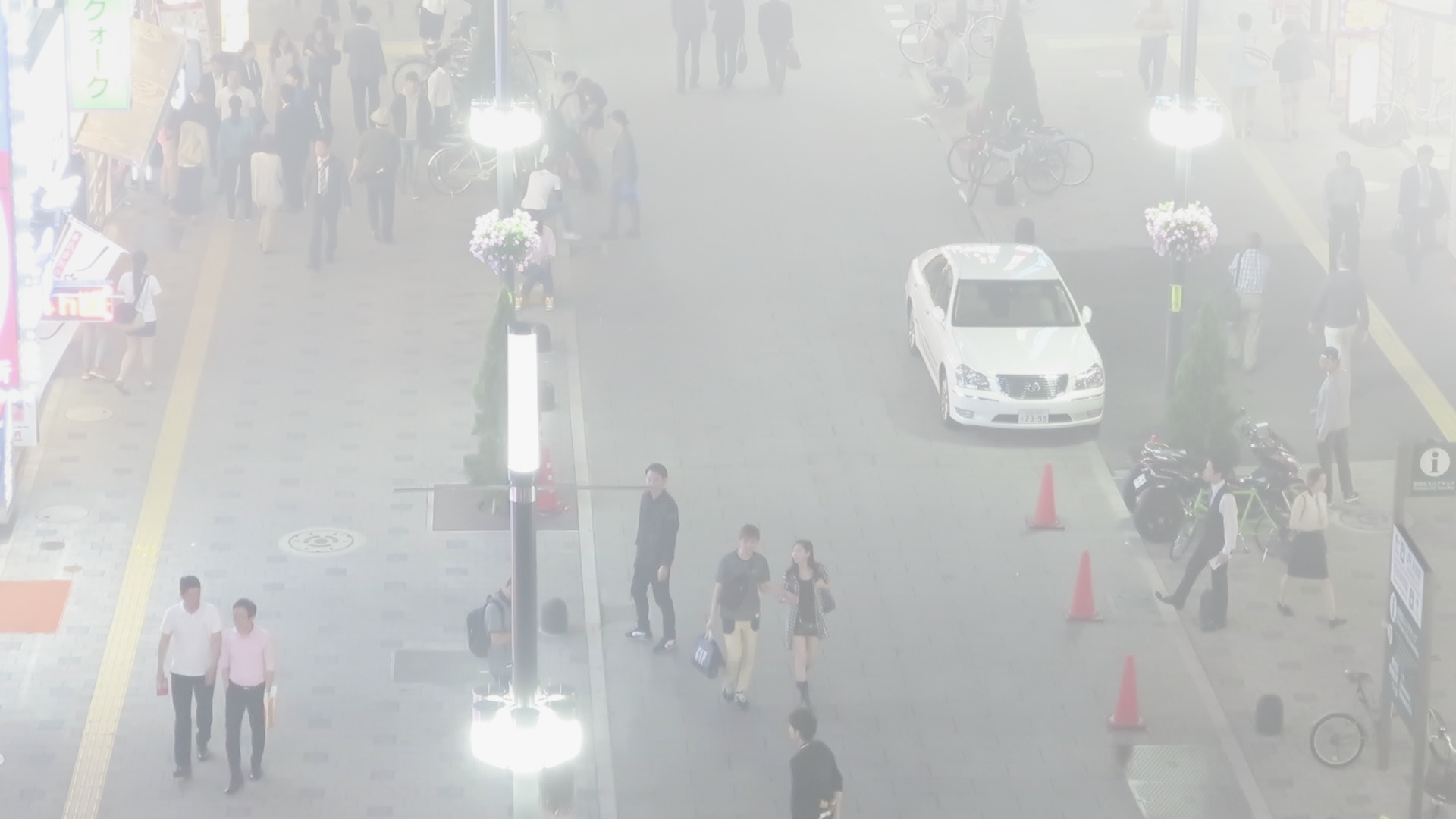}\\[-\baselineskip]}}
    &
    \subfigure[MOT17-11]{\bmvaHangBox{
        \includegraphics[width=0.25\linewidth]{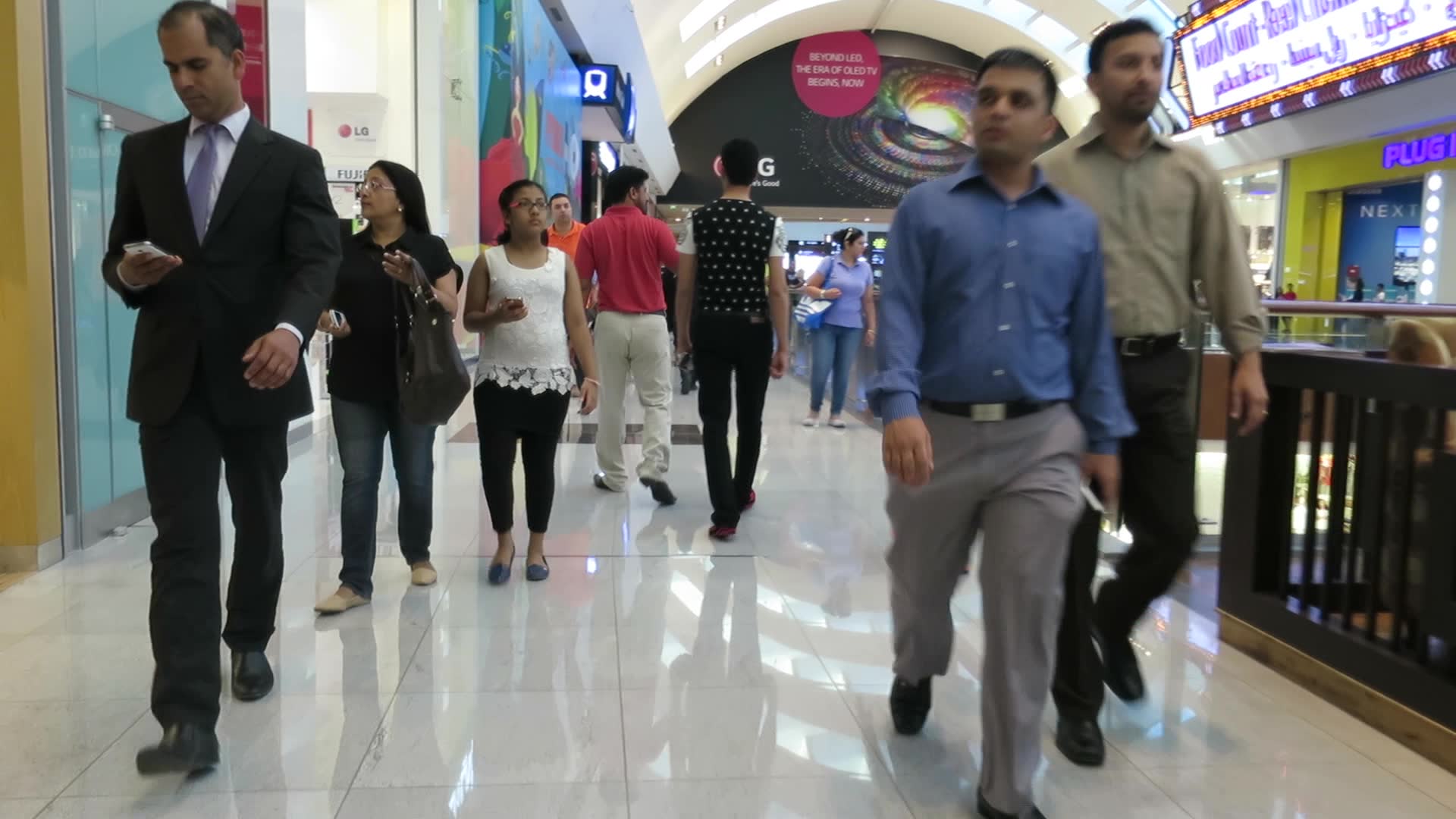}\\[-\baselineskip]
        \includegraphics[width=0.25\linewidth]{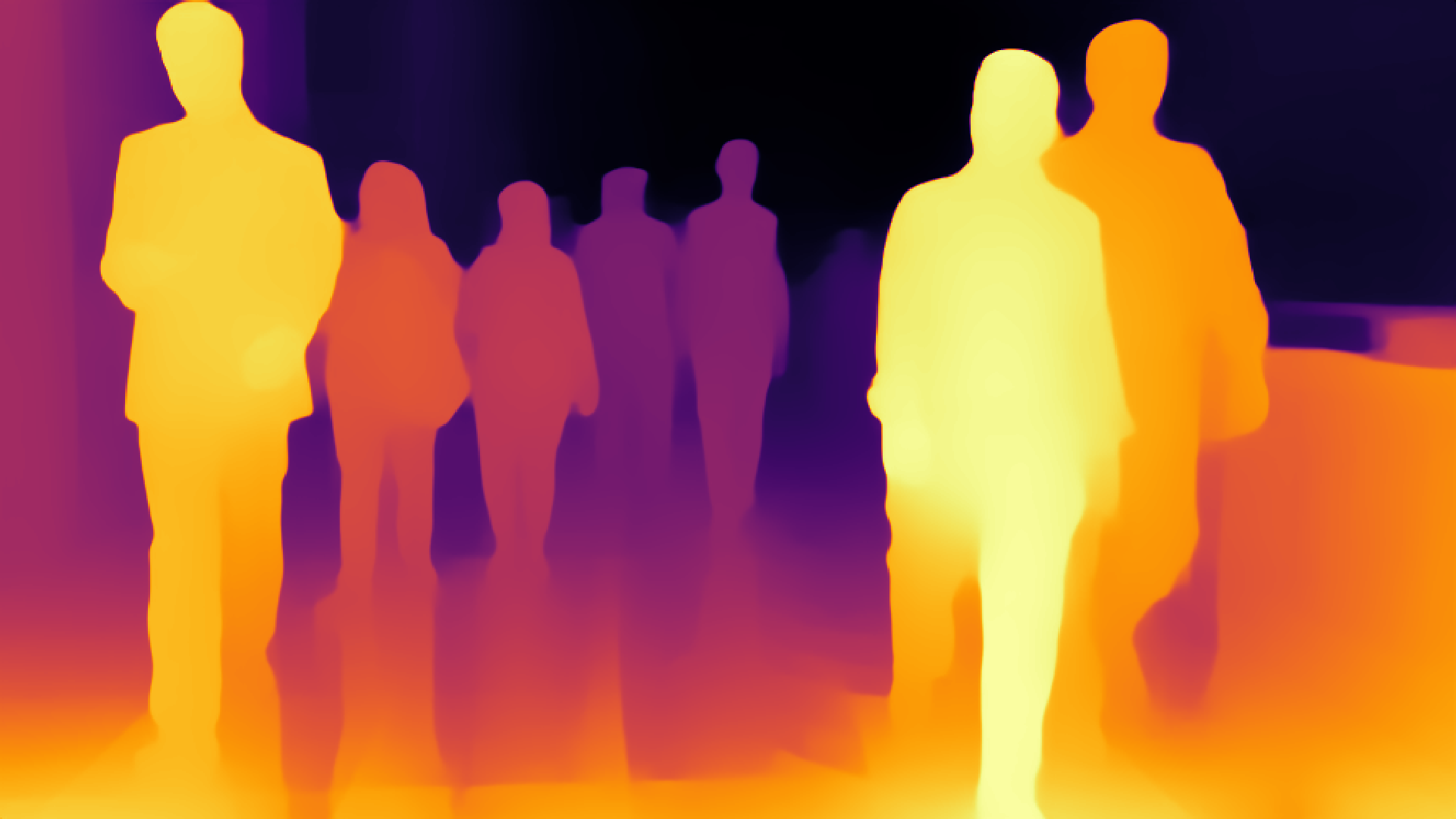}\\[-\baselineskip]
        \includegraphics[width=0.25\linewidth]{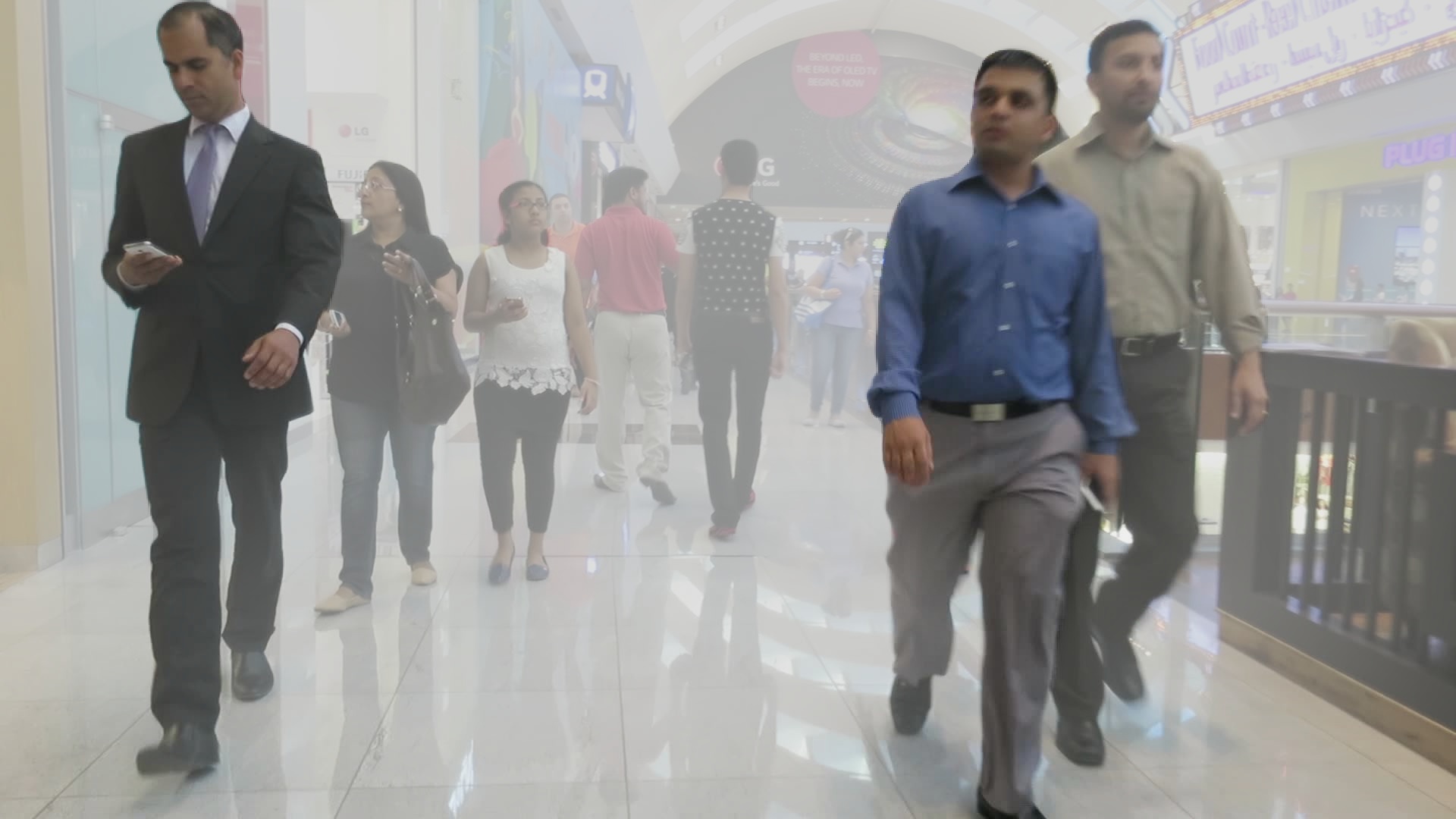}\\[-\baselineskip]}}
    &
    \subfigure[MOT17-13]{\bmvaHangBox{
        \includegraphics[width=0.25\linewidth]{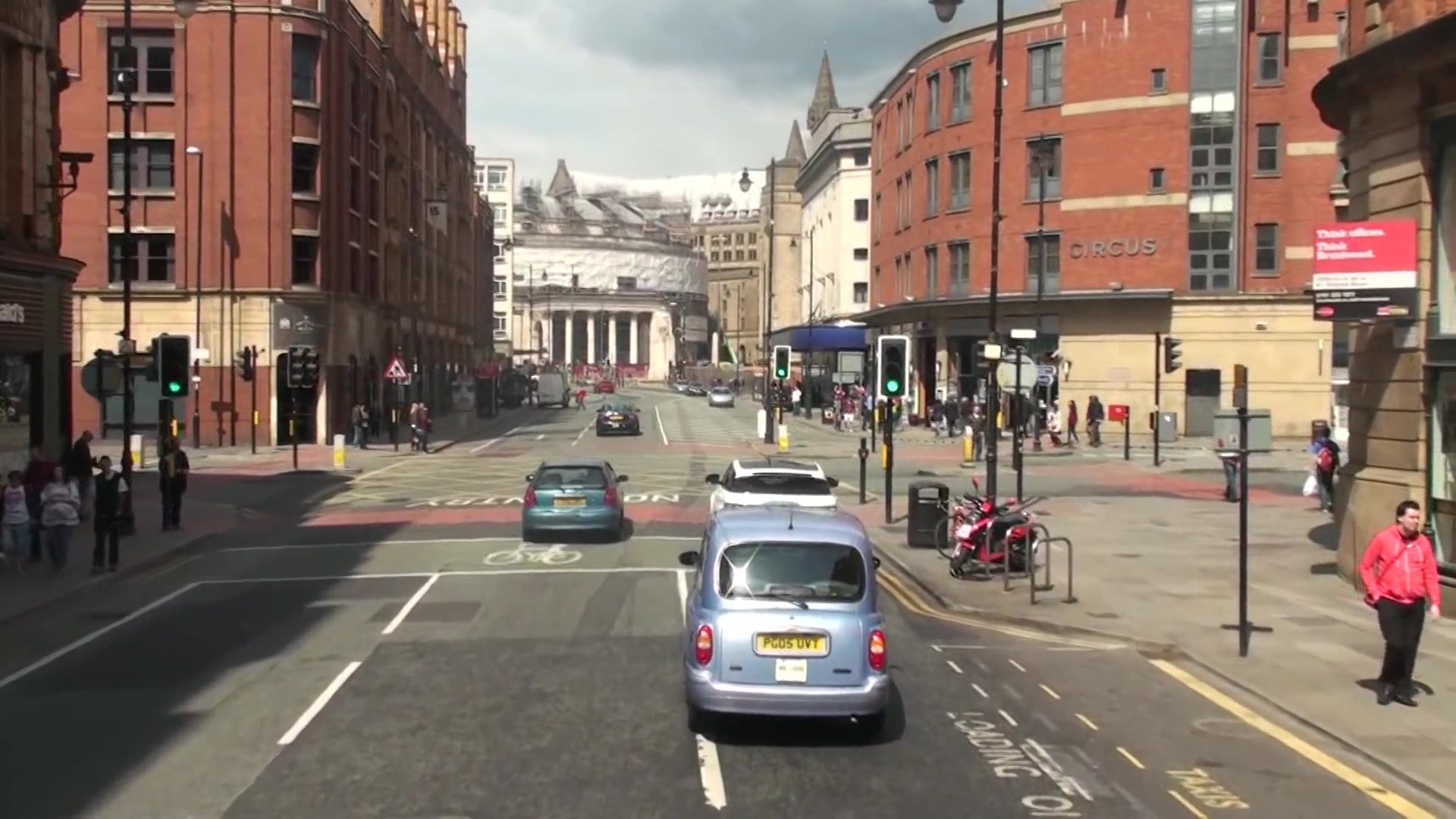}\\[-\baselineskip]
        \includegraphics[width=0.25\linewidth]{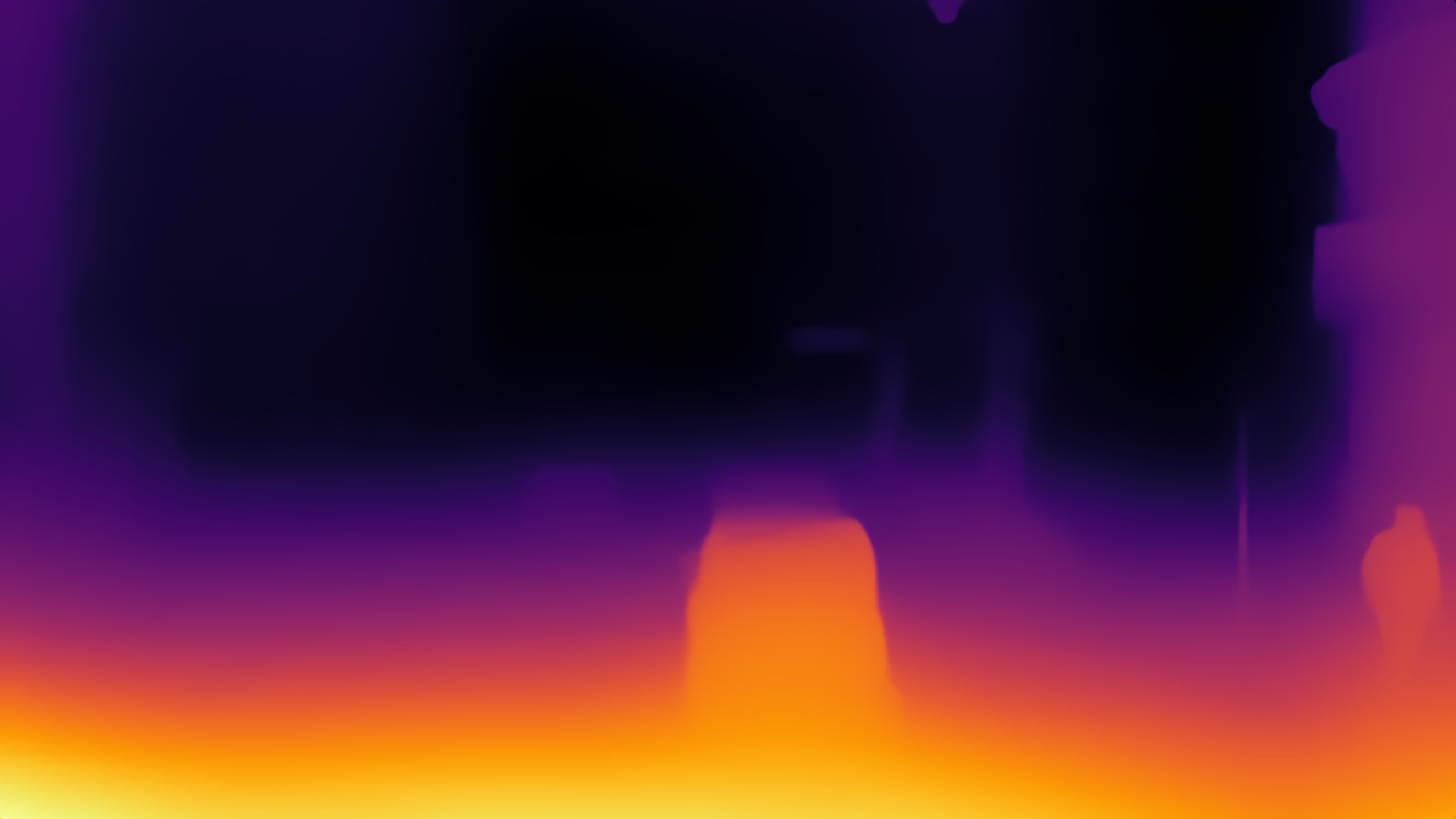}\\[-\baselineskip]
        \includegraphics[width=0.25\linewidth]{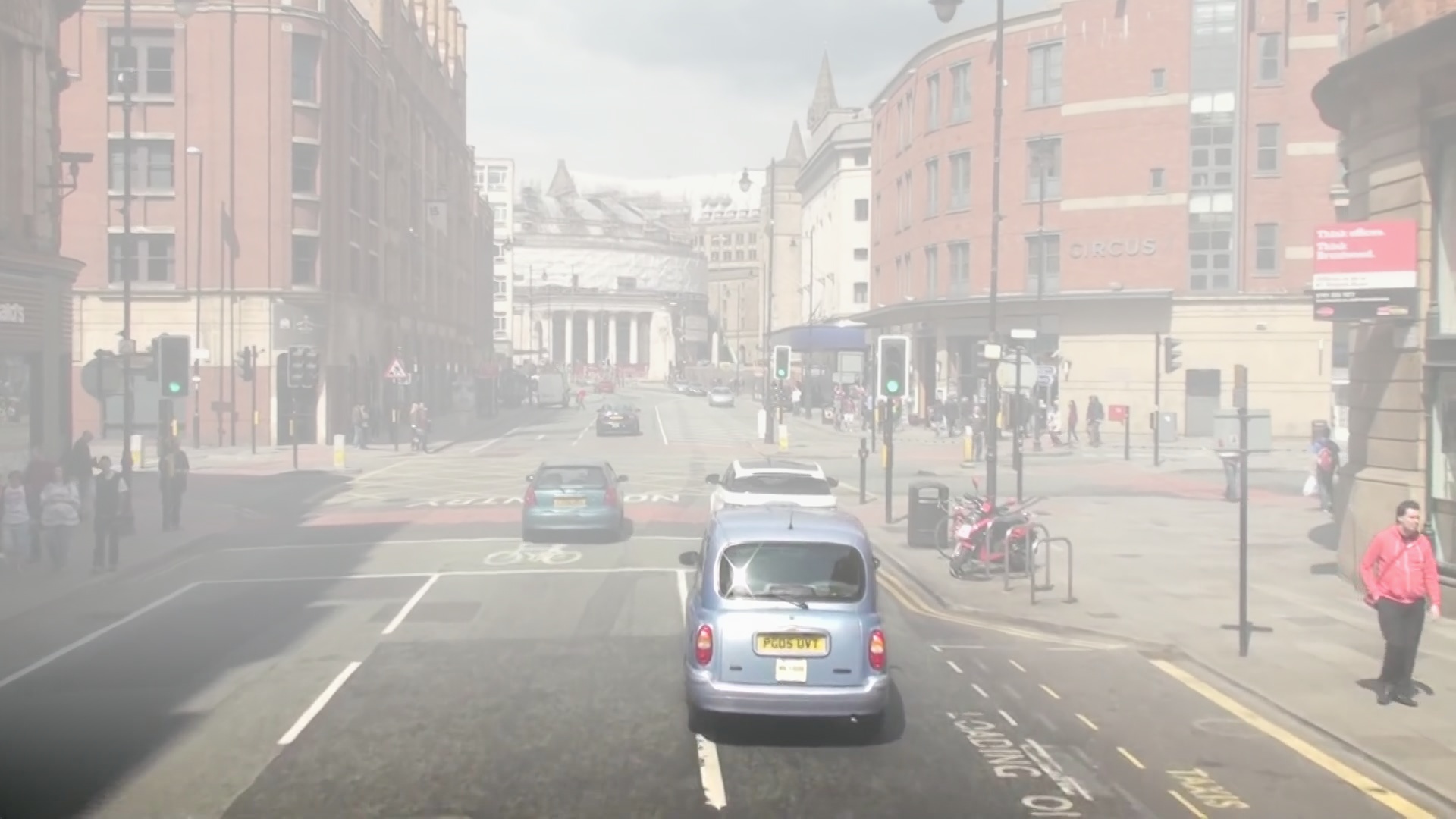}\\[-\baselineskip]}}
  \end{tabular}
  \vspace{-5pt}
  \caption{Illustration of fog (smoke) rendering in scenes from different domains: (a) daytime static frontal view; (b) nighttime surveillance view of a heavily illuminated, crowded square; (c) indoor crowd scene captured by a forward-moving camera; (d) daytime road view from a shaking, moving double-decker bus. From top to bottom: clear images, depth maps, foggy images.}
  \label{fig:mot17}
  \vspace{-5pt}
\end{figure*}

\subsection{Fog Formation Optical Model}
To understand the physics of vision in the atmosphere, we draw on the work of Narasimhan and Nayar~\cite{visionatmospher}, which consolidates fundamental research in this field, including contributions from Middleton~\cite{Middleton1951} and McCartney~\cite{McCartney}. 
We also  utilize Koschmieder's~\cite{koschmieder} mathematical model of haze and fog formation: 
\begin{equation}
    \mathbf{I(x)}=\mathbf{I_0(x)}\cdot \mathbf{T(x)} + L_\infty(1 - \mathbf{T(x)}),
    \qquad
    \mathbf{T(x)} = e^{-\beta\, \mathbf{D(x)}},
    \qquad
    \beta = \frac{-ln(0.05)}{V}, 
    \label{eq:opticalmodel}
\end{equation}
where $\mathbf{I(x)}$ is the foggy image, $\mathbf{I_{0}(x)}$ is the clear image, $\mathbf{T(x)}$ is the \textit{transmission map}, which defines the portion of object radiance reaching the camera after scattering in the atmosphere, and $L_\infty$ is the \textit{atmospheric light at the horizon}. 
The \textit{attenuation coefficient} $\beta$ is inversely proportional to the \textit{visibility} $V$ and regulates the fog intensity. 
The human eye contrast threshold $0.05$ is  proposed by Horvath~\cite{Horvath}. 
This model is commonly used in computer vision, computer graphics and dehazing approaches~\cite{tan, He, foggycityscape, leeaboutkoschmieder, 3dcommoncorruptions, fog4game, FRIDA}.
Examples of the $\beta$ calculation can be found in the supplementary material.

\subsection{Atmospheric Light}
Natural fog appearance varies significantly depending on lighting conditions, time of day, atmospheric pollutants, scene composition, and camera settings.
These variations lead to a wide range of fog hues and tones observed in photographs. 
In our fog formation model (see, Eq.~(\ref{eq:opticalmodel})), the \textit{atmospheric light at the horizon} $L_\infty$ regulates the fog color. 
If the sky is visible, we estimate it by averaging intensities of "far away" pixels on the estimated depth map. 
This aligns the fog color with the sky, ensuring consistency with the overall scene illumination.
However, in non-sky scenarios, such as surveillance settings where cameras point downward or in indoor scenes, the light at the horizon is not directly accessible. 
Moreover, highly illuminated night scenes pose significant difficulties.

Traditional methods, such as defining the image's brightest pixels as atmospheric light~\cite{tan} or using the image’s average intensity~\cite{3dcommoncorruptions}, can result in fog that is either too bright or unrealistically dark, especially in surveillance and indoor scenes (see the supplementary material).
This occurs because the brightest pixels often correspond to objects like white cars or white clothing, while the average illumination in non-sky images tends to be low.

To address these issues, we employ the \textit{dark channel prior} concept introduced by~He~\etal~\cite{He}, which provides a novel prior for image pixel intensity distribution based on statistical analysis of numerous non-sky haze-free images. The dark channel map $\mathbf{I}^{\text{dark}}\mathbf{(x)}$ of an image $\mathbf{I(x)}$ can be calculated from:
\begin{equation}
    \mathbf{I}^{\text{dark}}\mathbf{(x)} = \min_{\mathbf{y}\in\Omega\mathbf{(x)}}\left(\min_{c\in\{R,G,B\}} \mathbf{I}^{c}\mathbf{(y)}\right),
    \label{eq:dark}
\end{equation}
where $\Omega\mathbf{(x)}$ is a local patch centered at pixel $\mathbf{x}$ and $\mathbf{I}^{c}\mathbf{(y)}$ is the intensity across three color channel $c$ for each pixel $\mathbf{y}$ of the patch.
Constructing the dark channel by sliding a small window over the image and pooling each patch's brightness by its minimum value across the three color channels, we generate a new intensity distribution, the top 10\% brightest pixels of which describe the atmospheric light.
This method filters out excessively bright values and effectively thresholds dark pixels, resulting in a stable atmospheric light estimation and a more realistic fog appearance.
Detailed comparisons and visualizations of fog appearance using different methods across various scenarios are provided in the supplementary material.

\subsection{Heterogeneous Effect}
In nature, 
air turbulence can cause random variations in fog intensity across different spatial regions, leading to the formation of irregular cloud clusters in the atmosphere.
To better capture this diversity and the dynamic nature of fog, we simulate heterogeneous fog by incorporating a \textit{turbulence texture} $\mathbf{\tau(x)}$.
Following~\cite{Zdrojewska, fog4game}, we generate it by combining $N$ levels of pseudo-random gradient \textit{Perlin noise} $\mathbf{P_n(x)}$~\cite{perlin, perlinnumpy} with various frequencies and amplitudes. We then redefine the transmission map $\mathbf{T(x)}$ in Eq.~\eqref{eq:opticalmodel} by blending the depth map $\mathbf{D(x)}$ with the turbulence texture $\mathbf{\tau(x)}$:

\begin{equation}
    \mathbf{T(x)} = e^{-\beta\, \mathbf{\tau(x)}\cdot\mathbf{D(x)}},
    \quad
    \text{where}
    \quad
    \mathbf{\tau(x)} = \sum_{n=1}^{N}\frac{\mathbf{P_n(x)}}{2^{n}}.
    \label{eq:turbulence}\end{equation}
To maintain visual consistency and avoid flickering effects throughout a video, we use the same turbulence texture for an entire sequence. 
Since the videos in MOT datasets are typically short (only a few seconds), we can reasonably neglect the gradual movement of fog over time, which is usually slow.

Fig.~\ref{fig:mot17} demonstrates the results achieved with our fog simulation framework. For more detailed explanations and further visualizations, please refer to the supplementary material.


\subsection{Validating Fog Appearance}
To assess the perceptual quality of fog simulations and validate our approach, we conducted a user study where participants compared real foggy photographs, simulations from other methods, and our approach.
Viewers also rated pairs of augmented videos – with homogeneous and heterogeneous fog.
We recruited 48 participants with varying levels of experience in computer graphics and image processing. 
Given the subjectivity of human visual perception, we employed the Mean Opinion Score to compare and evaluate all methods.
Our approach, when evaluated on images, was judged to be $12\%$ more realistic for heterogeneous fog and $29\%$ more realistic for homogeneous fog compared to the state-of-the-art methods (see the supplementary materials). 
Remarkably, heterogeneous fog simulation in videos was perceived as more realistic than homogeneous, contrary to the findings in image augmentation.
The dynamic nature of videos, with their capture of motion and changes over time, provides a more immersive experience, helping viewers engage more deeply with the content.
This makes heterogeneous fog in videos more realistic, unlike to the uniform and regularized homogeneous fog typically seen in photographs. 
This finding highlights the importance of heterogeneous fog rendering in videos,  which, to our knowledge, has not been done before and actually makes our method more suitable for MOT tasks. 
Comprehensive validation results, including both quantitative and qualitative evaluations of fog appearance, are available in the supplementary materials.

\vspace{-2pt}
\section{MOT in Foggy Environments}
\vspace{-3pt}
Understanding the impact of fog on MOT performance and its limitations is crucial for further developing MOT robustness.
In this section, we provide a comprehensive evaluation of existing MOT methods on the MOT17 dataset augmented with various types of fog.  

\vspace{-5pt}
\subsection{Multiple Object Trackers}
We select five MOT methods representing different tracking concepts and paradigms to analyze their robustness to foggy conditions: ByteTrack~\cite{bytetrack}, Tracktor++~\cite{tracktor}, CenterTrack~\cite{centertrack}, FairMOT~\cite{FairMOT} and TransCenter~\cite{transcenter}.
These trackers show strong performance and SOTA results on established benchmarks consisting of clear atmospheric conditions.

\textit{ByteTrack}~\cite{bytetrack} follows the classical tracking-by-detection paradigm, similar to DeepSort~\cite{DeepSort}, and treats detection and tracking as separate steps. 
It employs a Kalman filter~\cite{kalman} for motion prediction and utilizes appearance features for re-ID and data association. 
To enhance DeepSort, the authors introduce two key improvements: 
Firstly, they handle all detections, regardless of their confidence scores, removing detector's thresholds and delegating detection filtering to the tracking stage. 
This enables the tracker to discover valuable detections, mitigating potential detector failures.  
Secondly, they propose an improved association method BYTE, to effectively handle object linking.  
Our work adopts the official ByteTrack implementation using the anchor-based YOLOX~\cite{yolox} detector.

\textit{Tracktor++}~\cite{tracktor} aims to challenge the prevailing tracking-by-detection paradigm by introducing a new concept of employing a detector for motion prediction.
By applying a regression head, the detector learns to propagate bounding boxes to the next frame, reducing the need for traditional motion prediction models, such as the Kalman filter, and facilitating the data association step. 
To address low-frame-rate videos, it applies the constant velocity assumption and utilizes re-ID with a Siamese neural network to handle long-term occlusion. 
Tracktor++ demonstrates strong performance regarding false negatives (FN) and identity preservation (IDF1) compared to other trackers. 
Our research employs the official Tracktor++ implementation, which is built on the Faster~R-CNN~\cite{fasterrcnn} detector with Feature Pyramid Network~\cite{FPN} and ResNet-101~\cite{resnet} backbone.

\textit{CenterTrack}~\cite{centertrack}, like Tracktor++, follows the idea of jointly learning detection and displacement over time, eliminating separate motion prediction and data association models, but adopts a point-based object representation, tracking their centers.
Built on the CenterNet~\cite{CenterNet} detector, CenterTrack propagates a heatmap of object keypoints across adjacent frames and predicts temporal point offsets using optical flow techniques.

\textit{FairMOT}~\cite{FairMOT} belongs to a class of trackers that jointly learn features for object detection and instance re-ID. 
Finding a feature balance for such two distinct tasks is challenging. 
Detection treats objects within a class as identical, focusing to distinguish between different classes, while re-ID needs to differentiate between instances of the same class. 
Simultaneous learning leads to a model bias toward object detection. 
FairMOT addresses this by employing two parallel heads on the ResNet34~\cite{resnet} feature extractor, using the DLA~\cite{DLA} multi-layer feature fusion model.
Simular to CenterTrack, it is based on the CenterNet detector, utilizing center heatmaps and tracking objects as points. 
Using a traditional Kalman filter~\cite{kalman} and appearance features for re-ID, FairMOT is capable to handle long-range associations and effectively address occlusion scenarios.

\textit{TransCenter}~\cite{transcenter} is a recent transformer-based MOT approach that utilizes object center heatmap representations. 
It employs powerful attention-based encoder and decoder, and integrates a Query Learning Network to generate tracking queries for producing object displacement vectors over time. 
The TransCenter's decoder jointly learns object detection and its displacements, resulting in higher performance. 
Built on the improved Pyramid Vision Transformer (PVT \textit{v2})~\cite{pvtv2} encoder, it overcomes efficiency issues present in other transformer-based trackers such as TrackFormer~\cite{trackformer} and TransTrack\cite{transtrack}, caused by the use of DETR~\cite{detr}. Differently to the DETR encoder, PVT is a completely convolution-free transformer backbone, which significantly reduces transformer sequence lengths and therefore computational cost. 
Moreover, unlike ViT~\cite{vit}, PVT is enhanced by a feature pyramid and copes well with the dense pixel-level structure, making it ideal for center heatmap predictions and yielding superior results compared to other trackers.

\subsection{Robustness Evaluations}

We augment multi-domain video scenarios of the MOTChallenge benchmark (third release: MOT17 dataset) with fog or smoke (for indoor scenes) of varying intensity levels, denoted as \textit{Fog~1}, \textit{Fog~2}, \textit{Fog~3}, and \textit{Fog~4} to categorize decreasing visibility as described in Sec.~\ref{seq:depthmap}.
We then assess MOT robustness to fog using
HOTA~\cite{HOTA}, CLEAR-MOT~\cite{CLEAR} and IDF1~\cite{IDF1} metrics. 
Each metric evaluates overall MOT performance but emphasizes different aspects. 
HOTA provides a balanced score for both detection and association, while traditional MOTA tends to prioritize detection accuracy over association. IDF1 focuses on the identity assignments, i.e. association. 
Since the MOTChallenge benchmark primarily focuses on pedestrian tracking, our analysis is reported for this particular object class.
We present the quantitative evaluation results in Tab.~\ref{tab:evalMOT}.

\begin{table*}[t]
\centering
\resizebox{\textwidth}{!}{\begin{tabular}{@{}cclccccc@{}}
\toprule
\multicolumn{2}{c}{Method} & Setup & HOTA$^\uparrow(\%)$ & MOTA$^\uparrow(\%)$ & MOTP$^\uparrow(\%)$ & IDF1$^\uparrow(\%)$ & IDsw$^\downarrow$ \\
\midrule
\multirow{25}{*}{
            \begin{sideways}
                Homogeneous Fog
            \end{sideways}}
    & \multirow{5}{*}{
        \begin{sideways}
            ByteTrack
        \end{sideways}}
        & Clear & 78.53 & 76.14 & 84.78 & 78.95 & 389\\
    &   & Fog 1 & 78.20	& 75.64	& 84.64	& 77.70 & 363 \\
    &   & Fog 2 & 74.74 & 72.65 & 84.72 & 74.78 & 324 \\
    &   & Fog 3 & \textbf{58.15} & \textbf{47.25} & 84.14 & \textbf{57.21} & 205\\
    &   & Fog 4 & \textbf{27.54} & \textbf{11.41} & 83.04 & \textbf{19.16} & 79 \\ 
\cmidrule{2-8}
    & \multirow{5}{*}{
        \begin{sideways}
            Tracktor++
        \end{sideways}}
        & Clear & 65.64 & 61.77 & \textbf{88.81} & 67.82 & \textbf{217} \\
    &   & Fog 1 & 56.19 & 52.28 & 86.44 & 55.43 & 287 \\
    &   & Fog 2 & 40.66 & 31.69 & 84.08 & 36.96 & 270 \\
    &   & Fog 3 & 18.79 & \phantom{0}8.19 & 79.76 & 12.05 & \phantom{0}33  \\
    &   & Fog 4 & \phantom{0}9.24 & \phantom{0}2.45 & 74.51 & \phantom{0}3.88 & \phantom{00}4 \\ 
\cmidrule{2-8}
    & \multirow{5}{*}{
        \begin{sideways}
            CenterTrack
        \end{sideways}}
        & Clear & 60.39 & 64.74 & 81.10 & 53.91 & 1849 \\
    &   & Fog 1 & 58.20 & 60.78 & 80.97 & 52.03 & 1826 \\
    &   & Fog 2 & 48.33 & 43.89 & 81.61 & 43.18 & 1736 \\
    &   & Fog 3 & 29.39 & 17.57 & \textbf{84.23} & 23.01 & \phantom{0}548 \\
    &   & Fog 4 & 11.51 & \phantom{0}3.68 & 82.99 & \phantom{0}5.68 & \phantom{0}187 \\
\cmidrule{2-8}
    & \multirow{5}{*}{
        \begin{sideways}
            FairMOT
        \end{sideways}}
        & Clear & 75.40 & 72.96 & 83.25 & 78.36 & 299 \\
    &   & Fog 1 & 73.38 & 69.13 & 83.40 & 75.35 & 264 \\
    &   & Fog 2  & 66.45 & 59.94 & 83.37 & 67.85 & 212 \\
    &   & Fog 3  & 39.18 & 24.88 & 81.65 & 35.42 & 148 \\
    &   & Fog 4  & 13.05 & \phantom{0}4.10 & \textbf{83.71} & \phantom{0}7.43 & \phantom{0}18 \\ 
\cmidrule{2-8}
    & \multirow{5}{*}{
        \begin{sideways}
           TransCenter
        \end{sideways}}
        & Clear & \textbf{84.66} & \textbf{85.53} & 87.19 & \textbf{83.86} & 280 \\
    &   & Fog 1 & \textbf{83.09} & \textbf{80.50} & \textbf{86.90} & \textbf{82.87} & 274 \\
    &   & Fog 2 & \textbf{78.68} & \textbf{74.62} & \textbf{85.28} & \textbf{78.42} & 257 \\
    &   & Fog 3 & 50.87 & 34.26 & 82.32 & 46.61 & 157 \\
    &   & Fog 4 & 26.66 & \phantom{0}6.26 & 77.96 & 12.42 & 120 \\ 
\bottomrule
\end{tabular}
\quad
\begin{tabular}{@{}clccccc@{}}
\toprule
& Setup & HOTA$^\uparrow(\%)$ & MOTA$^\uparrow(\%)$ & MOTP$^\uparrow(\%)$ & IDF1$^\uparrow(\%)$ & IDsw$^\downarrow$ \\
\midrule
\multirow{25}{*}{
            \begin{sideways}
                Heterogeneous Fog
            \end{sideways}}
         & Clear & 78.53 & 76.14 & 84.78 & 78.95 & 389 \\
         & Fog 1 & 78.42	& 76.06 & 84.41 & 78.40 & 371 \\
         & Fog 2 & 76.68	& 74.29	& 84.73	& 76.45 & 333 \\
         & Fog 3 & \textbf{69.35} & \textbf{61.98}	& \textbf{84.77}	& \textbf{69.69} & 261 \\
         & Fog 4 & \textbf{38.02} & \textbf{19.14} & \textbf{85.11} & \textbf{31.06} & 127 \\ 
\cmidrule{2-7}         
         & Clear & 65.64 & 61.77 & \textbf{88.81} & 67.82 & \textbf{217} \\
         & Fog 1 & 59.37	& 55.89	& 86.86	& 59.45 & 281\\
         & Fog 2  & 48.38 & 43.35 & 84.76 & 46.64 & 248 \\
         & Fog 3  & 26.34 & 13.57 & 82.47 & 19.68 & \phantom{0}58\\
         & Fog 4  & 10.10 & 2.39 & 76.54 & \phantom{0}3.74 & \phantom{0}16 \\ 
\cmidrule{2-7}
         & Clear & 60.39 & 64.74 & 81.10 & 53.91 & 1849 \\
         & Fog 1 & 59.32 & 62.35 & 80.98 & 53.04 & 1799 \\
         & Fog 2 & 51.46 & 50.6 & 80.76 & 45.47 & 1960 \\
         & Fog 3 & 37.34 & 26.27 & 83.82 & 30.78 & \phantom{0}888 \\
         & Fog 4 & 17.21 & \phantom{0}9.33 & 83.44 & 11.01 & \phantom{0}414 \\ 
\cmidrule{2-7}
         & Clear & 75.40 & 72.96 & 83.25 & 78.36 & 299\\
         & Fog 1 & 75.09 & 70.26 & 83.43 & 77.69 & 233\\
         & Fog 2 & 69.10 & 63.69 & 83.46 & 70.67 & 231\\
         & Fog 3 & 53.18 & 44.06 & 82.15 & 53.00 & 205\\
         & Fog 4 & 24.99 & 11.65 & 82.03 & 19.11 & \phantom{0}40 \\ 
\cmidrule{2-7}
         & Clear & \textbf{84.66} & \textbf{85.53} & 87.19 & \textbf{83.86} & 280\\
         & Fog 1 & \textbf{83.89} & \textbf{81.93} & \textbf{86.97} & \textbf{83.21} & 276\\
         & Fog 2 & \textbf{80.47} & \textbf{77.32} & \textbf{86.14} & \textbf{80.31} & 258\\
         & Fog 3 & 67.90 & 57.42 & 83.44 & 66.93 & 223\\
         & Fog 4 & 35.24 & 16.32 & 81.34 & 27.01 & 137 \\ 
\bottomrule
\end{tabular}}
\caption{MOT evaluation on the augmented MOT17 dataset with homogeneous (left) and heterogeneous (right) fog. Increasing fog intensity denoted from \textit{Fog 1} to \textit{Fog 4}. The best scores across all trackers in the same fog level are highlighted in bold.}%
\label{tab:evalMOT}
\vspace{-10pt}
\end{table*}

Despite demonstrating strong performance under clear atmospheric conditions, all trackers exhibit a significant drop in foggy scenarios.
Considering homogeneous fog, ByteTrack shows the highest robustness with slow performance degradation 
from \textit{Clear} 
to \textit{Fog~4} level, showing $-0.4\%, -5\%, -26\%, -65\%$ HOTA drops and $-0.7\%, -5\%, -38\%, -85\%$ MOTA drops.
These results may be attributed to its strategy of handling even uncertain detections regardless of their confidence score, allowing the tracker to perform well even in low visibility conditions, thereby correcting potential detector failures.
Despite having a slightly larger percentage decrease compared to the classical ByteTrack, the transformer-based TransCenter achieves the best tracking performance until \textit{Fog~2}.
FairMOT exhibits better robustness than CenterTrack, despite having a similar detector, possibly due to its re-ID branch and separate motion prediction and association models.
Although Tracktor++ and CenterTrack utilize the same tracking paradigm, the robustness of CenterTrack is better compared to Tracktor++, possibly because of its point-based object representation instead of an anchor-based one.
Tracktor++ shows the weakest performance with significant drops already at the beginning, degrading by $-10\%, -38\%, -71\%, -86\%$ (HOTA) and $-15\%, -49\%, -87\%, -96\%$ (MOTA), respectively. Detailed degradation scores for all trackers can be found in the supplementary material.

MOTA scores degrade worse than HOTA, indicating the crucial role of detection for robust tracking.
MOTP scores, which evaluate the localization precision, consistently yield high results unaffected by fog. 
The IDsw metric, indicating the number of identity switches during tracking, correlates with the total number of detected instances and decreases accordingly across all trackers due to the increased number of missed tracks. 

MOT approaches perform slightly better under heterogeneous fog, likely due to transparent regions between clouds of thicker fog. A detailed analysis is provided in the supplementary material.
Nevertheless, while some trackers perform better in reduced visibility than others, a huge gap persists compared to their clear performance. 
Fig.~\ref{fig:main} already illustrated that despite the human eye's ability to track objects in extremely severe fog, current MOT methods fail.
A key takeaway is the importance of selecting the right tracking architecture to cope with fog. 
As observed, a successful strategy involves mitigating detector failures by processing even uncertain detections while improving an association model. 
Choosing a pixel-wise object center heatmap representation over anchor-based also yields better results.
Incorporating the prediction of temporal object displacement into detectors instead of using a separate model is still under-explored in terms of MOT robustness, while
transition from convolution-based architectures to vision transformers drastically enhances MOT.


\vspace{-5pt}
\section{Conclusion}
\vspace{-5pt}
By introducing a photorealistic fog augmentation method for arbitrary MOT datasets, we bridge the data gap for further enhancement of MOT robustness.
Our evaluations shed light on the limitations of various MOT paradigms and architectures when faced with foggy conditions and set the foundation for developing MOT methods capable to handle adverse atmospheric conditions.
Our augmentation method can be applied on any tracking dataset, enabling our community to investigate and improve the robustness of MOT approaches.


\paragraph*{Acknowledgments}
This work was partially funded by the Austrian Research Promotion Agency (FFG) under the project SAFER (894164).


\bibliography{egbib}
\end{document}